\documentclass{ecai} 

\usepackage{latexsym}
\usepackage{amssymb}
\usepackage{amsmath}
\usepackage{amsthm}
\usepackage{booktabs}
\usepackage{enumitem}
\usepackage{graphicx}
\usepackage{color}
\usepackage{xcolor} 
\usepackage{colortbl}
\usepackage{algorithm}
\usepackage{algorithmic}
\usepackage{appendix}
\usepackage{multirow}
\usepackage{multicol}
\usepackage{subcaption}

\newtheorem{theorem}{Theorem}
\newtheorem{lemma}{Lemma}
\newtheorem{corollary}{Conclusion}

\newtheorem{definition}{Definition}

\newcommand{\BibTeX}{B\kern-.05em{\sc i\kern-.025em b}\kern-.08em\TeX}

\newlength\myindent
\newcommand\bindent{
  \begingroup
  \setlength{\itemindent}{\myindent}
  \addtolength{\algorithmicindent}{\myindent}
}
\newcommand\eindent{\endgroup}

\begin{document}

\begin{frontmatter}

\paperid{1405} 

\title{SelfBC: Self Behavior Cloning for Offline \\ Reinforcement Learning}


\author[A]{\fnms{Shirong}~\snm{Liu}}
\author[B]{\fnms{Chenjia}~\snm{Bai}}
\author[A]{\fnms{Zixian}~\snm{Guo}}
\author[A]{\fnms{Hao}~\snm{Zhang}}
\author[C]{\fnms{Gaurav}~\snm{Sharma}}
\author[A]{\fnms{Yang}~\snm{Liu}\thanks{Corresponding Author. Email: liuyang@hit.edu.cn}}

\address[A]{Harbin Institute of Technology, Harbin, China} 
\address[B]{Shanghai AI Laboratory, Shanghai, China}
\address[C]{University of Rochester, Rochester, USA} 

\begin{abstract}
Policy constraint methods in offline reinforcement learning employ additional regularization techniques to constrain the discrepancy between the learned policy and the offline dataset. 
However, these methods tend to result in overly conservative policies that resemble the behavior policy, thus limiting their performance. We investigate this limitation and attribute it to the static nature of traditional constraints. In this paper, we propose a novel dynamic policy constraint that restricts the learned policy on the samples generated by the exponential moving average of previously learned policies. 
By integrating this self-constraint mechanism into off-policy methods, our method facilitates the learning of non-conservative policies while avoiding policy collapse in the offline setting.
Theoretical results show that our approach results in a nearly monotonically improved reference policy.
Extensive experiments on the D4RL MuJoCo domain demonstrate that our proposed method achieves state-of-the-art performance among the policy constraint methods.
\end{abstract}

\end{frontmatter}

\section{Introduction}
Reinforcement learning (RL) \citep{sutton2018rlintroduction} has made great progress in recent years, driven by breakthroughs such as deep Q-learning \citep{mnih2015dqn} and policy gradient methods \citep{schulman2017ppo}. However, these progressions rely heavily on massive online interactions with the environment, which can be impractical, inefficient, and even risky in real-world scenarios such as autonomous driving \citep{sallab2017autodriving} and healthcare \citep{yu2021healthcare}. Therefore, the ability to learn a policy from existing datasets becomes crucial for practical applications, highlighting the importance of offline reinforcement learning (offline RL) \citep{levine2020offline}.

Traditional off-policy RL methods often encounter challenges such as value overestimation and policy collapse when applied directly in offline scenarios. These issues arise from the distribution shift between the learned policy and the offline dataset; the learned Q function may suffer from extrapolation errors and overestimate the value of actions that are not observed in the offline dataset, further prompting the learned policy to produce out-of-distribution (OOD) actions. To address these challenges, current offline RL methods integrate regularization techniques to prevent the production of OOD actions. 
One popular way of regularization is to use policy constraints that 
explicitly restrict the discrepancy between the learned policy and the offline dataset. 
Conventional policy constraint methods mainly explore different divergence metrics to restrict the distribution shift, such as behavior cloning (BC) \citep{fujimoto2021td3bc,fakoor2021CDC_sac+bc}, dataset support \citep{wu2022spot, zhang2021brac+,peng2019awr, ashvin2020awac}, maximum mean discrepancy \citep{kumar2019BEAR_MMD}, and Fisher divergence \citep{kostrikov2021fisher-brc}. However, due to the static nature of the reference policy, these methods often suffer from overly tight constraints and learn overly conservative policies.

To mitigate this problem, some works aim to relax the constraint within the policy constraint framework. TD3+BC has been augmented by adding a policy-refinement phase \citep{beeson2022improvingtd3bc}, where, after the initial training process is finished, the weightage of the BC term is reduced.
ReBRAC \citep{tarasov2024rebrac} adds extra design elements and hyperparameter tuning to relax the constraint and improve performance. 
These methods loosen the policy constraint by introducing additional complexity while 
keeping the reference policy fixed to the dataset behavior policy. 
Because offline datasets often contain samples with low quality, constraining on these samples results in suboptimal learned policies. To reduce the impact of low quality samples, weighted behavior cloning (wBC) \citep{chen2020bail, wang2020crr, peng2023wPC} methods distinguish the quality of dataset samples by assigning weights, and the weighted behavior model priors \citep{siegel2020abm} are learned by weighted likelihood, which are further used to constrain the learned policy. However, these methods often entail intricate weight design heuristics and require the training of additional value functions for weight computation. 
Beyond the policy constraint framework, many studies leverage model-based rollouts \citep{yu2020mopo, kidambi2020morel, yu2021combo} or uncertainty estimation \citep{wu2021uwac, bai2022pbrl, an2021edac, nikulin2023sac-rnd} for value regularization to handle the over conservatism.

In order to overcome the limitations of policy constraint methods mentioned above, we introduce a novel self-constraint framework for offline RL.
This framework constrains the learned policy on a dynamic reference policy, which is the exponential moving average (EMA) of previously learned policies.
In the learning process, we first initialize the reference policy using a pre-trained policy from an existing policy constraint method, and then progressively update the reference policy toward the learned policy. The learned policy is optimized by combining our proposed policy constraint and existing off-policy algorithms. 
Our approach offers several advantages.
First, our policy constraint is dynamically updated. This dynamic approach serves to alleviate policy conservatism and facilitates the learning of non-conservative policies in a stable manner. 
Second, since the learned policy is constrained to be close to the reference policy, the increment in distribution shift of the newly updated reference policy will be minimal. 
Lastly, the update process of our reference policy can be seen as a conservative policy update process \citep{kakade2002cpi}, which our theoretical results demonstrate can achieve nearly monotonically improved reference policies. 
Experimental results demonstrate that our dynamic reference policy approach leads to progressively improved learned policies during training. Our method achieves state-of-the-art performance on the D4RL \citep{fu2020d4rl} MuJoCo benchmark, particularly in non-expert datasets. 

Our main contributions are as follows:
\vspace{-0.5em}
\begin{enumerate}
    \item We empirically demonstrate the problem of policy conservatism in traditional policy constraint methods that adopts behavior cloning and attribute it to the static nature of the reference policy.
    \item We propose Self Behavior Cloning (SelfBC) as a novel dynamic policy constraint. By integrating SelfBC into an off-policy algorithm, our method is able to relax the overly conservative learned policy while achieving a stable policy improvement process and avoiding policy collapse.
    \item We extend the analysis of conservative policy iteration to the offline setting and theoretically establish that our approach results in a nearly monotonically improved reference policy.
    \item Extensive experiments on the offline datasets from the D4RL MuJoCo domain demonstrate that our proposed algorithm achieves state-of-the-art performance among policy constraint methods.
\end{enumerate}

\vspace*{-0.2in}
\section{Related Works}
\paragraph{TD3+BC with relaxed constraint}
TD3+BC with policy refinement \citep{beeson2022improvingtd3bc} relaxes the constraint by adding a policy refinement step after the original TD3+BC training is completed, where the weightage of the BC term is reduced during the refinement step. ReBRAC \citep{tarasov2024rebrac} improves performance by relaxing the BC constraint of TD3+BC and integrating several design elements such as larger batches, layer norm for the critic networks, critic penalty, deep networks, and hyperparameter tuning. 
We focus on dynamic reference policies instead.

\paragraph{Weighted Behavior Cloning Methods}
Weighted behavior cloning (wBC) methods aim to address the limitation of BC. As BC is a form of imitation learning, which requires high-quality datasets that perform similar to human experts, wBC can have some sample selective effect on the imperfect datasets. In particular, methods that employ the policy constraint with reverse-KL regularization can derive an advantage-weighted behavior policy as the optimal solution \citep{wang2018marwil,peng2019awr,ashvin2020awac}.
One line of wBC methods contains three parts: learning a value function for advantage estimation, transforming advantages to weights for dataset samples, and learning a policy to perform weighted behavior cloning. 
For instance,
BAIL \citep{chen2020bail} learns a value function that evaluates the upper envelope of dataset actions to select high-quality actions from the dataset, and 
CRR \citep{wang2020crr} learns a value function that evaluates
the current learned policy and designs identifier and exponential functions to compute weights from advantages. 
wPC \citep{peng2023wPC}
integrates the weighted behavior cloning loss into the TD3+BC framework. 
ABM \citep{siegel2020abm} 
learns a policy prior via advantage-weighted behavior cloning to formulate the reverse KL constraint. Our method directly updates the reference policy towards current learned policy instead of constructing the reference policy by weighting dataset samples.

\paragraph{Conservative Policy Iteration}
Conservative Policy Iteration (CPI) \citep{kakade2002cpi} guarantees monotonic improvement by incrementally updating the learned policy as stochastic mixtures of consecutive policies. CPI has been integrated into deep reinforcement learning \citep{vieillard2020deepcpi}. 
CPI can be extended to general stochastic policy classes by directly constraining the discrepancy of consecutive policies, such as TRPO \citep{schulman2015trpo} and PPO \citep{schulman2017ppo}. 
Behavior Proximal Policy Optimization \citep{zhuang2023behaviorppo} (BPPO) can improve a previously learned policy by learning from the offline dataset. The constraint of BPPO is the KL divergence between consecutive policies, which is similar to the online PPO. CPI-RE \citep{hu2023cpi_re} iteratively refines the reference policy as the current learned policy, and uses both the reference policy and the dataset to formulate the constraint. By applying CPI, our method slowly updates the reference policy towards current learned policy by EMA.

\vspace*{-0.1in}
\section{Preliminaries}
\subsection{Offline Reinforcement Learning}
We consider the standard RL formulation~\cite{sutton2018rlintroduction} in terms of a Markov Decision Process (MDP) defined as $(\mathcal{S}, \mathcal{A}, \mathcal{P}, r, \gamma, \rho_0)$, where $\mathcal{S}$ is the state space, $\mathcal{A}$ is the action space, $\mathcal{P}:\mathcal{S} \times \mathcal{A} \rightarrow \Delta(S)$ is the transition function, where $\Delta(S)$ is the probability simplex over $\mathcal{S}$, $r:\mathcal{S} \times \mathcal{A} \rightarrow \mathbb R$ is the reward function, $\gamma$ is the discount factor, and $\rho_0 \in \Delta{\mathcal{S}}$ is the initial state distribution. RL aims to find the optimal policy $\pi(a|s)$ that maximizes the expected discounted return $J\left(\pi\right) = \mathbb{E}_\pi\left[\sum_{t=0}^{\infty} \gamma^t r\left(s_t,a_t\right)\right]$. The Q function and the value function of a policy $\pi$ are defined as $Q_\pi\left(s,a\right)=\mathbb{E}_\pi\left[\sum_{t=0}^{\infty} \gamma^t r\left(s_t,a_t\right)|s_0=s, a_0=a\right]$ and $V_\pi\left(s\right)=\mathbb{E}_\pi\left[\sum_{t=0}^{\infty} \gamma^t r\left(s_t,a_t\right)|s_0=s\right]$, respectively. And the advantage function is $A_\pi \left(s,a\right)=Q_\pi \left(s,a\right)-V_\pi \left(s\right)$. In the deep RL setting, $\pi_{\theta}$ and $Q_{\phi}$ are used to denote policy and Q networks, respectively, with $\theta$ and $\phi$ representing corresponding network parameters.

In offline RL, the agent cannot collect samples by interacting with the environment; instead, it must learn the policy from a previously collected dataset $\mathcal{D}=\left\{\left(s_i,a_i,r_i,s_i' \right)\right\}_{i=1}^N$, where the policy used to collect the dataset is unknown and can be single or mixed. Due to the difficulty of policy evaluation without online sample collection, offline RL is challenging.

\vspace*{-0.1in}
\subsection{TD3+BC}
TD3+BC is a minimalist solution to offline RL, based on TD3, it only adds a behavior cloning (BC) term to the policy objective and balances the BC term and the original Q term by a hyperparameter $\alpha$. 
TD3+BC maintains two learned Q networks $Q_{\phi_1}, Q_{\phi_2}$, a learned policy network $\pi_\theta$ and their target networks $Q_{\bar \phi_1}, Q_{\bar \phi_2}, \pi_{\bar \theta}$. 
The policy optimization objective of TD3+BC is:
\begin{eqnarray}\label{eq:TD3+BC}
\mathcal{L}_{\pi}^{\text{TD3+BC}}\left(\theta\right)
= \mathbb{E}_{\left(s, a\right) \sim \mathcal{D}}
\left[
    \alpha Q_{\phi_1}\left(s, \pi_\theta\left(s\right)\right) 
    - \left(\pi_{\theta}\left(s\right)-a\right)^2
\right]
\end{eqnarray}

To unify the hyperparameter settings in different offline datasets, TD3+BC additionally normalizes the Q term in Eq.~\eqref{eq:TD3+BC}. 

TD3+BC learns the Q networks in the same way as TD3:
\begin{eqnarray}\label{eq:Qloss_TD3+BC}
\mathcal{L}_{Q}^{\textnormal{TD3+BC}}\left(\phi_i\right) = \mathbb{E}_{\left(s, a, r, s'\right) \sim \mathcal{D}}\left[(Q_{\phi_i}\left(s, a\right) - y)^2\right], i=1,2
\end{eqnarray}
where $y$ is the target and its computation incorporates techniques of clip double q-learning and target policy smooth regularization \citep{fujimoto2018td3}:
\begin{eqnarray}
\nonumber y &=& r + \gamma \min_{i=1,2} Q_{\bar{\phi}_i}\left(s',a'\right) \\
a' &=& \pi_{\bar \theta}\left(s'\right) + \epsilon, \quad \epsilon \sim clip\left(\mathbb{N}\left(0,\sigma\right), -c, c\right)
\end{eqnarray}
where $\sigma$ is the action noise scale and $c$ is the clip range.

\begin{figure*}[htbp]
    \centering
        \begin{subfigure}{.46\linewidth}
            \setlength{\abovecaptionskip}{-1pt}
            \setlength{\belowcaptionskip}{2pt}
            \includegraphics[width=1\linewidth]{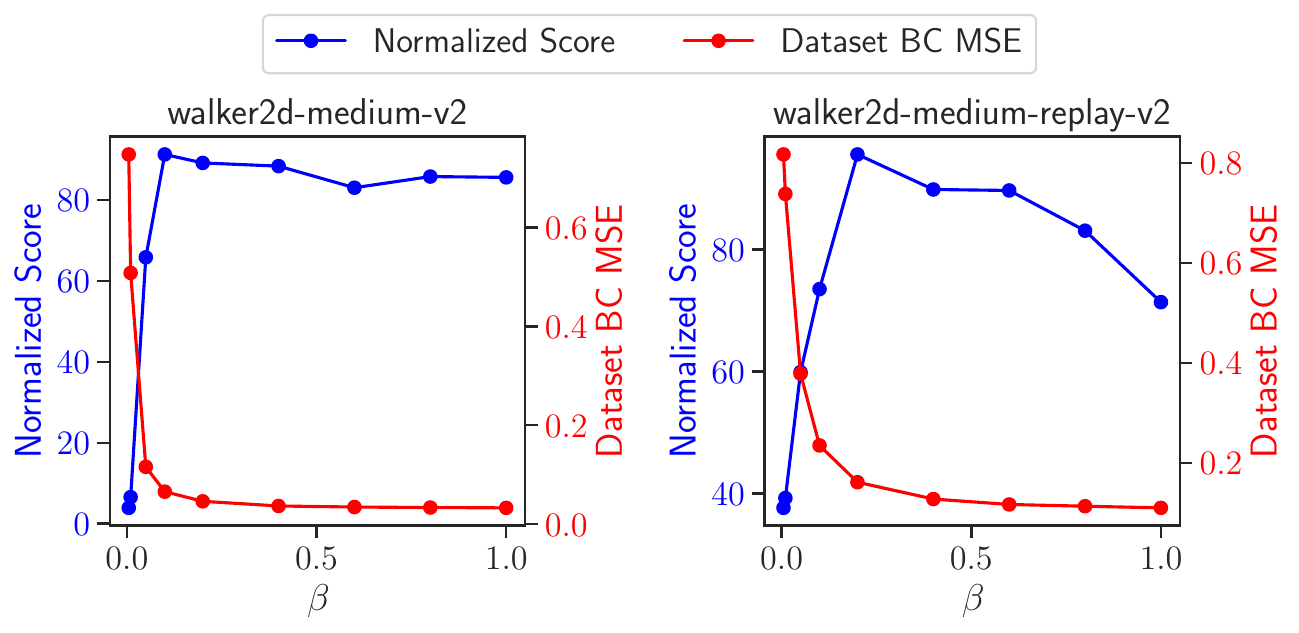}
            \caption{}
            \label{fig:analys2a}
        \end{subfigure}
        \begin{subfigure}{.46\linewidth}
            \setlength{\abovecaptionskip}{-1pt}
            \setlength{\belowcaptionskip}{2pt}
            \includegraphics[width=1\linewidth]{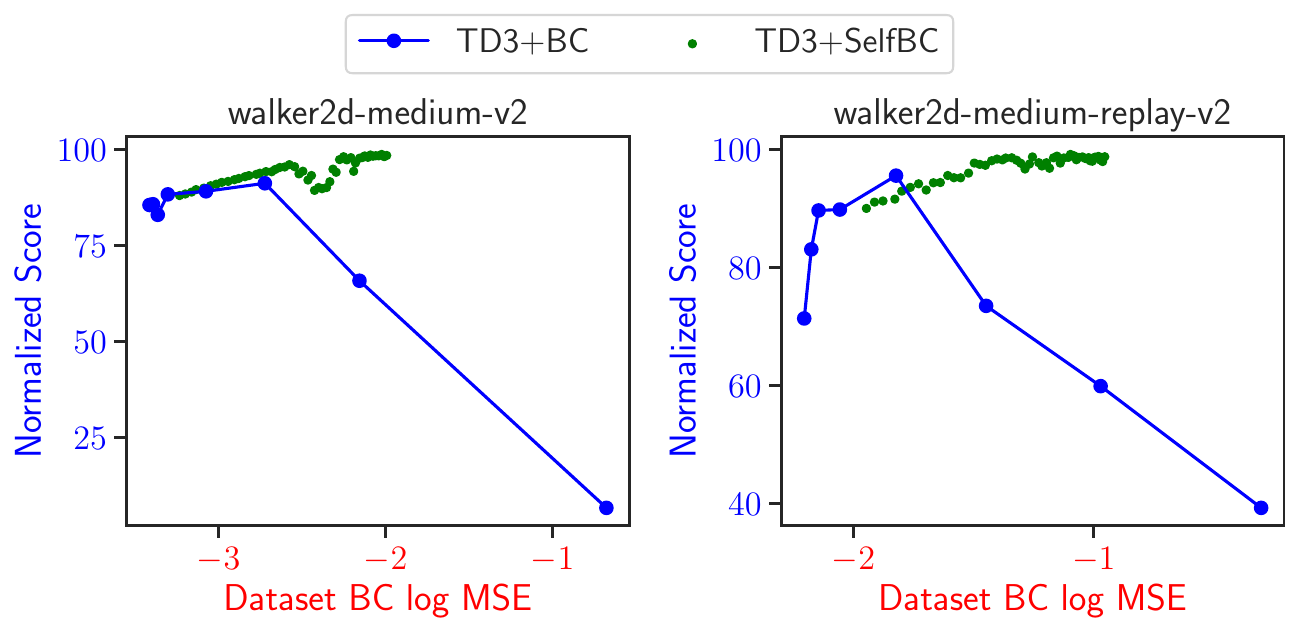}
            \caption{}
            \label{fig:analys2b}
        \end{subfigure}
    \setlength{\belowcaptionskip}{1pt}
    \caption{
      For different values of $\beta$, TD3+BC was trained on each of the D4RL \citep{fu2020d4rl} datasets and, for the final learned policy, values were recorded for the normalized score and dataset BC MSE,  i.e., the average BC penalty in Eq.~\eqref{eq:TD3+BC}. For two datasets, we illustrate in (a) the dependence of the normalized score and dataset BC MSE on $\beta$, and in (b) we plot the normalized score against the dataset BC log MSE obtained for different $\beta$ in blue (for TD3+BC) and compare against the corresponding results obtained with our proposed TD3+SelfBC under an identical training setting, which are plotted in green. (The results are averaged across multiple seeds).
    }
    \label{fig:analys2}
\end{figure*}

\vspace*{-0.1in}
\section{Analysis of Over Conservatism}
\label{sec:td3bc_analysis}
In this section, we examine the issue of over-conservatism
and analyze why it persists within the policy constraint framework.

To examine the influence of the
policy constraint on the asymptotic performance, we add a hyperparameter $\beta$ into Eq.~\eqref{eq:TD3+BC}, yielding
\begin{eqnarray}
\mathbb{E}_{\left(s, a\right) \sim \mathcal{D}}
\left[
    Q_{\phi_1}\left(s, \pi_\theta\left(s\right)\right) - \beta\left(\pi_\theta\left(s\right)-a\right)^2
  \right]
  \label{eq:betaaddedTD3nBBObjective}
\end{eqnarray}
where the coefficient $\alpha$ is dropped since we do not change it.

We train TD3+BC on different $\beta$ values and illustrate the experimental results on two offline datasets in Fig. \ref{fig:analys2}, where $\beta=1.0$ corresponds to the original version of TD3+BC. (Fig. \ref{fig:td3bc_beta_score_mse} of Appendix \ref{Appendix:full_results_alalysis} provides results for all datasets).
In Fig. \ref{fig:analys2a}, we illustrate the dependence on $\beta$ of the normalized score and the dataset BC mean squared error (MSE), i.e., the average BC penalty in Eq.~\eqref{eq:TD3+BC}, of the final learned policy. The dataset BC MSE quantifies the distribution shift of the learned policy from the offline dataset. 
It can be seen that as $\beta$ decreases from $1$, i.e., the policy constraint relaxes, the distribution shift (dataset BC MSE) of the final learned policy consistently increases, and the performance (normalized score) initially improves but subsequently declines. 
Fig. \ref{fig:analys2b} plots the normalized score vs. dataset BC log MSE of the learned policy obtained for different $\beta$ values for TD3+BC in blue; a trace of sample values for the corresponding quantities obtained during the training process for our proposed TD3+SelfBC algorithm under an identical training setting is shown in green (see Appendix~\ref{Appendix:full_results_alalysis} for additional detail).
We can see that as the distribution shift of the learned policy increases, the performance of TD3+BC does not always increase.

From the preceding observations, we conclude that:
\begin{itemize}
    \vspace{-0.6em}
    \item Tight policy constraints can stabilize policy optimization and ensure quick convergence. However, the asymptotic performance tends to be suboptimal due to the overly conservative nature of the learned policy, which closely resembles the behavior policy.
    \item Although relaxing the policy constraint may improve the potential theoretical optimal policy, in practice, it destabilizes the training process and often leads to policy collapse.
    \vspace{-0.6em}
\end{itemize}

It is obvious that the performance of TD3+BC can be improved by adjusting the value of $\beta$ which balances the risk of policy collapse and over conservatism. However, manually adjusting the degree of policy constraint on each target task is cumbersome.
Our proposed dynamic policy constraint for offline RL can achieve stable policy improvement without the need for intricate hyperparameter tuning.

\section{Proposed Method}

\subsection{General Behavior Cloning Framework}

We first introduce the general behavior cloning framework, which learns a deterministic policy $\pi_\theta$ to clone the samples produced by a general reference policy $\pi_{\textnormal{ref}}$. Then we demonstrate that behavior-cloning-based approaches used in existing policy constraint methods to learn a deterministic policy \cite{fujimoto2021td3bc, chen2020bail, wang2020crr, peng2023wPC} are special cases of cloning a general reference policy.

We start by introducing the weighted behavior cloning objective:
\begin{eqnarray}\label{eq:wbc}
\mathbb{E}_{s \sim \mathcal{D}, a \sim \pi_b^{\mathcal{D}}\left(\cdot|s\right)}
\left[
    w(s,a) \cdot \left(\pi_\theta\left(s\right) - a\right)^2
\right]
\end{eqnarray}
where $\pi_b^{\mathcal{D}}$ denotes the ground truth behavior policy of the offline dataset $\mathcal{D}$, and $w(s,a)$ is the weight function.
This formulation involves an expectation over $\left(s,a\right)$ pairs. By moving the expectation over $a$ inside the behavior cloning term, we obtain the following equivalent objective with the expectation solely over $s$ (a simplified proof is provided below, full proof is provided in Appendix \ref{appendix:equivalent_wbc}):
\begin{eqnarray}\label{eq:general_wbc}
\mathbb{E}_{s \sim \mathcal{D}}
\left[
    \left(
        \pi_\theta(s) - 
        \mathbb{E}_{a \sim q_w\left(\cdot|s\right)}\left[a\right]
    \right)^2
\right]
\end{eqnarray}
where $q_w\left(a|s\right) = \frac{\pi_b^{\mathcal{D}}\left(a|s\right) \cdot w(s,a)}{\int \pi_b^{\mathcal{D}}\left(a|s\right) \cdot w\left(s,a\right) da}$ represents the normalized weighted behavior policy of the dataset.
\begin{proof}
Considering given a state $s$,
\begin{eqnarray}
    &&\left(\pi_\theta(s) - \mathbb{E}_{a \sim q_w\left(\cdot|s\right)}\left[a\right]\right)^2 
    = \left(\mathbb{E}_{a \sim q_w\left(\cdot|s\right)}\left[\pi_\theta\left(s\right) - a\right]\right)^2
    \nonumber \\
    &=& \mathbb{E}_{a \sim q_w\left(\cdot|s\right)}\left[\left(\pi_\theta\left(s\right) - a\right)^2\right] - {\mathrm{Var}}_{a \sim q_w\left(\cdot|s\right)}\left[a\right] \nonumber \\
    &\equiv& \mathbb{E}_{a \sim \pi_b^{\mathcal{D}}(\cdot|s)}\left[w(s,a) \cdot \left(\pi_\theta\left(s\right) - a\right)^2\right]
\end{eqnarray}
where the variance in the second equation is independent of $\pi_\theta$.
\end{proof}

Now we consider two  particular cases of Eq.~\eqref{eq:general_wbc}.
\paragraph{Expected Behavior Cloning (EBC)}
First, note that when we set $w(s,a)=1$ for all $(s,a) \in \mathcal{D}$, we get an equivalent of the naive behavior cloning objective, given by
\begin{eqnarray}\label{eq:TD3+BC_equivalent}
\mathbb{E}_{s \sim \mathcal{D}}\left[(\pi_\theta\left(s\right)- \mathbb{E}_{a \sim \pi_b^{\mathcal{D}}\left(\cdot|s\right)}\left[a\right])^2\right]
\end{eqnarray}
To implement Eq.~\eqref{eq:TD3+BC_equivalent}, we do not need to precisely model $\pi_b^{\mathcal{D}}\left(\cdot|s\right)$, as it is often multi-modal and complex. Instead, we learn the expected action $\mathbb{E}_{a \sim \pi_b^{\mathcal{D}}\left(\cdot|s\right)}\left[a\right]$, which is deterministic given a state $s$ and can be easily learned by behavior cloning. 
We can learn a neural network-modeled behavior policy $\pi_b$ through behavior cloning:
\begin{eqnarray}\label{eq:pib}
\pi_b = \mathop{\arg\min}_\pi \mathbb{E}_{\left(s,a\right) \in \mathcal{D}}\left[\left(\pi_b\left(s\right) - a\right)^2\right]
\end{eqnarray}
And Eq.~\eqref{eq:TD3+BC_equivalent} can be rewritten as:
\begin{eqnarray}\label{eq:EBC}
\mathcal{L}_{\pi}^{\textnormal{EBC}}\left(\theta\right)=\mathbb{E}_{s \sim \mathcal{D}}\left[\left(\pi_\theta\left(s\right)-\pi_b\left(s\right)\right)^2\right]
\end{eqnarray}
As $\pi_b(s)$ approximates the expected action $\mathbb{E}_{a \sim \pi_b^D\left(\cdot|s\right)}\left[a\right]$, we refer to $\mathcal{L}_{\pi}^{\textnormal{EBC}}\left(\theta\right)$ as the Expected Behavior Cloning (EBC) objective. 
Here, $\pi_b$ is a special case of reference policy restricting $\pi_{\theta}$.

\paragraph{Expected Weighted Behavior Cloning (EWBC)}
Second, note that when $w(s,a)$ is an arbitrary function, the reference policy corresponding to Eq.~\eqref{eq:general_wbc} is $q_w$, which is a special reference policy corresponding to $w$ and $\pi_b^{\mathcal{D}}$. We call such a reference policy, $q_w$, the expected weighted behavior policy.
This indicates that the weighted behavior cloning method in Eq.~\eqref{eq:wbc} is actually a special case of modifying the reference policy by designing a special weight function.

\paragraph{General Behavior Cloning}
We introduce the general behavior cloning objective that involves a free-design reference policy $\pi_{\textnormal{ref}}$ as
\begin{eqnarray}\label{eq:general_bc}
\mathcal{L}_{\pi}^{\textnormal{{General BC}}}\left(\theta\right) = 
\mathbb{E}_{s \sim \mathcal{D}}\left[\left(\pi_\theta\left(s\right) - \mathbb{E}_{a \sim \pi_{\textnormal{ref}}\left(\cdot|s\right)}\left[a\right]\right)^2\right]
\end{eqnarray}

This objective covers the two cases mentioned above $(\pi_{\textnormal{ref}} = \pi_b$ for EBC, $\pi_{\textnormal{ref}} = q_w$ for EWBC). Both EBC and EWBC clone reference policies that derive from static behavior of offline dataset. 
We propose that, in the case of general behavior cloning, it is not necessary to design the reference policy that explicitly follows the static behavior of offline dataset, which may lead to suboptimal performance as shown in Section 4. 
Rather, the design of $\pi_{\textnormal{ref}}$ can be flexible as long as it maintains a low distribution shift from the offline dataset behavior, ensuring the preservation of valuable in-distribution information from the offline dataset. 
We introduce our proposed reference policy design in the next section.

\vspace*{-0.15in}
\subsection{Self Behavior Cloning}\label{sec:selfbc}
\vspace*{-0.1in}

We first introduce our design of the reference policy for Eq.~\eqref{eq:general_bc}, and then integrate our policy constraint into an off-policy algorithm. The underlying logic of our reference policy design is as follows:
\begin{itemize}
\vspace{-0.6em}
    \item The reference policy should be dynamically updated. Based on Section \ref{sec:td3bc_analysis}, the static reference policy is not suitable for the policy search with a large distribution shift, which restricts the algorithm performance.
    \item The performance of the reference policy should surpass the offline dataset and it would be advantageous if the performance of the reference policy is monotonically increased.
\vspace{-0.6em}
\end{itemize}

To this end, we propose the Self Behavior Cloning (SelfBC) objective. In the general BC objective in Eq.~\eqref{eq:general_bc}, our approach employs a novel reference policy $\pi_{\tilde \theta}$, which is the EMA of previously learned policies, yielding the SelfBC objective
\begin{eqnarray}
\mathcal{L}_{\pi}^{\textnormal{SelfBC}}\left(\theta\right) = 
  \mathbb{E}_{s \sim \mathcal{D}}\left[\left(\pi_\theta\left(s\right) - \pi_{\tilde \theta}\left(s\right)\right)^2\right]
\label{eq:selfbc}  
\end{eqnarray}
Note that $\pi_{\tilde \theta}$ is deterministic here. To progressively increase the performance of such reference policy, we integrate our SelfBC into the off-policy algorithm TD3 and obtain our TD3+SelfBC objective:
\begin{align}\label{eq:TD3+SelfBC}
\mathcal{L}_{\pi}^{\textnormal{TD3+SelfBC}}\!\left(\theta\right)
\!=\!\mathbb{E}_{s \sim \mathcal{D}}\!
\left[
    \alpha Q\left(s, \pi_\theta\left(s\right)\right) \! - \! \left(\pi_\theta\left(s\right) \!- \! \pi_{\tilde \theta}\left(s\right)\right)^2
\right]
\end{align}
where the reference policy $\pi_{\tilde \theta}$ is progressively updated towards the latest learned policy $\pi_\theta$ by EMA:
\begin{eqnarray}\label{eq:ref_EMA}
 \tilde\theta = \tau_{\textnormal{ref}} \, \theta + \left(1 - \tau_{\textnormal{ref}}\right) \tilde\theta 
\end{eqnarray}
where $\tau_{\textnormal{ref}}$ represents the soft update ratio, set to be near zero.

\begin{algorithm}[H]
  \caption{TD3+SelfBC and TD3+EBC}
  \label{alg:TD3+SelfBC}
\begin{algorithmic}
  \STATE {\bfseries Hyperparameters:} training steps $N_{\textnormal{SelfBC}}, N_{\textnormal{BC}}, N_{\textnormal{EBC}}$, soft update ratio $\tau, \tau_{\textnormal{ref}}$, coefficient $\alpha$.
    
  \STATE {\bfseries Initialize:} Q networks $Q_{\phi_1}, Q_{\phi_2}$ and target Q networks $Q_{\bar\phi_1} Q_{\bar \phi_2}$, policy network $\pi_\theta$ and target policy network $\pi_{\bar \theta}$

  \STATE {\bfseries TD3+EBC Part:} // learn a behavior policy
  \bindent
  \FOR{$t=1$ {\bfseries to} $N_{\textnormal{BC}}$}
    \STATE Sample mini-batch state-action pairs $\left(s, a\right)$ from $\mathcal{D}$
    \STATE Learn $\pi_b$ by the objective in Eq.~\eqref{eq:pib}
  \ENDFOR
  \eindent

  \STATE {\bfseries TD3+SelfBC Part:}
  \STATE \quad // load pre-trained results
  \STATE \quad Load pre-trained Q networks $Q_{\phi_1}, Q_{\phi_2}$ and policy networks $\pi_\theta$
  
  \STATE \quad // training preparations
  \STATE \quad Update target networks: $\bar{\phi}_{1,2} \leftarrow \phi_{1,2}$, $\bar{\theta} \leftarrow \theta$
  \STATE \quad Update reference policy network: ${\tilde \theta} \leftarrow \theta$
  
  \STATE // training
  \FOR{$t=1$ {\bfseries to} $N_{\textnormal{SelfBC}}$ or $N_{\textnormal{EBC}}$}
    \STATE Sample mini-batch transitions $\left(s, a, r, s^{\prime}\right)$ from $\mathcal{D}$
    \STATE Update $\phi_{1,2}$ by minimizing $\mathcal{L}_Q(\phi)$ in Eq.~\eqref{eq:Qloss_TD3+BC}
    \STATE {\bfseries For TD3+EBC:}
    \STATE \quad Update $\theta$ by maximizing $\mathcal{L}_{\pi}^{\textnormal{TD3+EBC}}(\theta)=$ \\ \quad\quad
    $\mathbb{E}_{s \sim \mathcal{D}}\left[\alpha Q\left(s, \pi_\theta\left(s\right)\right) - \left(\pi_\theta\left(s\right)-\pi_b\left(s\right)\right)^2\right]$
    \STATE {\bfseries For TD3+SelfBC:}
    \STATE \quad Update $\theta$ by maximizing $\mathcal{L}_{\pi}^{\textnormal{TD3+SelfBC}}(\theta)$ in Eq.~\eqref{eq:TD3+SelfBC}
    \STATE \quad Update reference policy network: \\ \quad\quad
    ${\tilde\theta} \leftarrow \tau_{\textnormal{ref}} \theta + \left(1 - \tau_{\textnormal{ref}}\right) \tilde\theta$
    \STATE Update target networks: \\ \quad
    $\bar{\phi}_{1,2} \leftarrow \tau \phi_{1,2} + \left(1-\tau\right) \bar{\phi}_{1,2}$, $\bar{\theta} \leftarrow \tau \theta + \left(1-\tau\right) \bar{\theta}$
  \ENDFOR
\end{algorithmic}
\end{algorithm}

Thus, our reference policy $\pi_{\textnormal{ref}}$ has both the advantages of dynamically updating and performance changing with the learned policy.
The reference policy updated by EMA
can be seen as maintaining an exponential average of the recently learned policies, which can further reduce the noise from the learned policy and result in better training dynamics \citep{morales-brotons2024ema_in_dl}.

We use the word ``self'' in our proposed objective \eqref{eq:TD3+SelfBC} because it is similar to self-distillation \citep{caron2021dino, he2020moco}, which learns a student model to simultaneously optimize the original objective and minimize the discrepancy between the student model and the teacher model, where the teacher model is updated towards the student model by EMA. In Eq.~\eqref{eq:TD3+SelfBC}, $\pi_\theta$ and $\pi_{\tilde \theta}$ respectively correspond to the student model and the teacher model of self-distillation.

For the practical implementation of TD3+SelfBC, $\pi_{\tilde \theta}$ should be properly initialized so that its performance is not worse than the behavior policy $\pi_{b}$. 
Random initialization cannot be employed because as a random policy will not contain any information about the dataset support. To stabilize the training process, we initialize $\pi_{\tilde \theta}$ using a pre-trained policy trained by an existing policy constraint method; we use TD3+EBC as the default pretraining method, which can also be BC and TD3+BC, we show details of these in Section \ref{sec:ablation}. As a pretraining algorithm, TD3+EBC integrates the EBC in Eq.~\eqref{eq:EBC} into TD3, which is provided in Algorithm~\ref{alg:TD3+SelfBC}. We first train a behavior policy $\pi_b$ for $N_{\textnormal{BC}}$ timesteps and then use it to train TD3+EBC for $N_{\textnormal{EBC}}$ timesteps to obtain a nearly converged Q network and policy network. 

Algorithm~\ref{alg:TD3+SelfBC} summarizes the complete flow of TD3+SelfBC and TD3+EBC. Besides the reference policy, we also
initialize the learned Q networks and the learned policy by the pre-trained results. We initialize all the target networks by the learned networks. Then, we train TD3+SelfBC for $N_{\textnormal{SelfBC}}$ timesteps and update $\pi_{\tilde \theta}$ towards the learned policy 
$\pi_\theta$ per policy update via EMA. The update process of $\pi_{\tilde \theta}$ is similar to the target policy network, but with a slower update ratio. 

\vspace*{-0.1in}
\subsection{Ensemble Reference Policies}

Noise in the reference policy updating process can destabilize the reference policy and lead to policy collapse. We, therefore, proposed to mitigate the effect of the noise using an ensemble training approach that we outline here. The full details are in Appendix \ref{appendix:td3+esbc}.

In the proposed TD3+SelfBC, the reference policy is influenced by two factors:
the pre-trained policy and the learned policy. 
The former initializes the reference policy, while the latter updates it. However, both factors may inject noise into the reference policy, even though the EMA process may reduce the impact of the noise from the learned policy.
To reduce the noise, we propose to
simultaneously train an ensemble of TD3+SelfBC trainers.
Note that although each trainer maintains its own Q networks, we do not ensemble all the Q networks of all trainers for uncertainty estimation as existing uncertainty-based methods for offline RL \citep{bai2022pbrl, an2021edac}. Most parts of each trainer are independent of other trainers, such as Q learning, target network updating, and reference policy updating, except the policy objective uses a shared reference action for all trainers. 
Specifically, the shared reference action is computed by averaging the reference actions from all trainers. 
Assume that the number of ensemble is $N_{\textnormal{ens}}$ and ensemble learned policies and ensemble reference policies are $\{\pi_\theta^i, \pi_{\tilde \theta}^i | i=1,2,...,N_{\textnormal{ens}}\}$. Given a state $s$, the shared reference action $\tilde a$ is defined as:
\begin{eqnarray}
    \tilde a = \frac{1}{N_{\textnormal{ens}}} \sum_{i=1}^{N_{\textnormal{ens}}} \pi_{\tilde \theta}^i\left(s\right)
\end{eqnarray}
We train each individual trainer separately using the same $\tilde a$:
\begin{eqnarray}
\mathcal{L}_{\pi}^{\textnormal{TD3+ESBC}}(\theta^i)=\mathbb{E}_{s \sim \mathcal{D}}\left[\alpha Q^i\left(s, \pi_{\theta}^i\left(s\right)\right) - \left(\pi_{\theta}^i\left(s\right) - \tilde a\right)^2\right]
\label{eq:TD3+ESBC}
\end{eqnarray}

\vspace*{-0.15in}
\subsection{Theoretical Analysis}\label{sec:theory}
By combining the conservative policy iteration (CPI) \citep{kakade2002cpi, vieillard2020deepcpi} and the behavior proximal policy optimization (BPPO) \citep{zhuang2023behaviorppo} we develop theoretical analysis for TD3+SelfBC that can be seen as an extension of the CPI to the offline setting. In our  analysis, which closely follows~\cite{zhuang2023behaviorppo}, we consider the stochastic policy case; the deterministic policy can be seen as a special stochastic policy with near-zero variance. We first formulate our dynamic reference policy as CPI. Then we discuss the performance bound of the reference policy in the offline setting. Finally, we show that the reference policy in our TD3+SelfBC can achieve nearly monotonic improvement.

\begin{definition}\label{thm:def_pi}
We denote $\pi_k$ as the reference policy of the k-th iteration, $\pi'$ as the new learned policy, and $\pi$ as the new reference policy. 
We update the reference policy by CPI:
\begin{eqnarray}\label{eq:CPI}
\pi\left(\cdot | s\right) = \left(1-\kappa\right) \pi_k\left(\cdot | s\right)+ \kappa \pi'\left(\cdot | s\right)
\end{eqnarray}
where hyperparameter $\kappa$ controls the conservative update process.
\end{definition}

\begin{lemma}\citep{kakade2002cpi, zhuang2023behaviorppo}\label{thm:performance_difference_arbitrary}
  The difference in expected discounted return between two arbitrary policies $\pi',\pi$ is
\begin{eqnarray}
J_\Delta\left(\pi',\pi\right) 
= J(\pi') - J(\pi) 
= \mathbb{E}_{s \sim \rho_{\pi'}\left(\cdot\right), a \sim \pi'\left(\cdot \mid s\right)}\left[A_\pi\left(s, a\right)\right]
\end{eqnarray}
where $\rho_\pi\left(s\right) = \sum_{t=0}^{\infty}\gamma^t P\left(s_t=s|\pi\right)$ represents the unnormalized discounted state visitation frequencies.
\end{lemma}

We then define the performance difference of two consecutive reference policies using Lemma \ref{thm:performance_difference_arbitrary}, we also approximate the difference on the offline dataset:
\begin{definition}\label{thm:def_J_delta}
The performance difference of consecutive reference policies $J_\Delta\left(\pi,\pi_k\right)$, and its approximation $\widehat{J}_{\Delta}\left(\pi, \pi_k\right)$ on offline dataset $\mathcal{D}$ are defined as:
\begin{eqnarray}
J_{\Delta}\left(\pi, \pi_k\right) = \mathbb{E}_{s \sim \rho_{\pi}(\cdot), a \sim \pi(\cdot \mid s)}\left[A_{\pi_k}\left(s, a\right)\right] \\
\widehat{J}_{\Delta}\left(\pi, \pi_k\right) = \mathbb{E}_{s \sim \rho_{\mathcal{D}}(\cdot), a \sim \pi\left(\cdot \mid s\right)}\left[A_{\pi_k}\left(s, a\right)\right]
\end{eqnarray}
\end{definition}

We state performance bounds for our conservatively updated reference policy as the following theorem, whose proof is deferred to Appendix \ref{app:proof_of_thm_bppo_for_ref} due to space limitations.
\begin{theorem}\label{thm:bppo_for_ref}
  The performance difference between two consecutive reference policies influenced by the offline approximation satisfies
\begin{eqnarray}
J_{\Delta}\left(\pi, \pi_k\right)
&\geq&
\widehat{J}_{\Delta}\left(\pi, \pi_k\right) 
\nonumber \\
&-& 
\frac{2\gamma\kappa^2}{1-\gamma}
\mathbb{A}_{\pi',\pi_k}
\mathbb{E}_{s\sim\rho_{\pi_k}} \left[ D_{TV}\left(\pi' \| \pi_k\right)\left[s\right] \right]
\nonumber \\
&-& 
\frac{2\gamma\kappa}{1-\gamma}
\mathbb{A}_{\pi',\pi_k}
\mathbb{E}_{s\sim\rho_{\pi_b}} \left[ D_{TV}\left(\pi_k \| \pi_b\right)\left[s\right] \right] 
\nonumber \\
&-& 
\frac{\gamma\kappa}{1-\gamma}
\mathbb{A}_{\pi',\pi_k}
\mathbb{E}_{s\sim\rho_{\mathcal{D}}} \left[ 1-\pi_b\left(a|s\right) \right] \label{eq:DRImprovementLowerBound}
\end{eqnarray}
where $\pi, \pi_k, \pi'$ are as in Definition \ref{thm:def_pi},
\begin{eqnarray}
\mathbb{A}_{\pi',\pi_k} = 2 \max_{s,a} \left| A_{\pi_k}\left(s,a\right)  \right| \max_s D_{TV}\left(\pi' \| \pi_k\right)\left[s\right]
\end{eqnarray}
and,
\begin{eqnarray}\label{eq:extended_offline_performance_difference}
\widehat{J}_{\Delta}\left(\pi, \pi_k\right)
&=& \mathbb{E}_{s \sim \rho_{\mathcal D}(\cdot), a \sim \pi(\cdot \mid s)}\left[A_{\pi_k}(s, a)\right] \nonumber \\
&=& (1-\kappa) \mathbb{E}_{s \sim \rho_{\mathcal D}(\cdot), a \sim \pi_k (\cdot \mid s)}\left[A_{\pi_k}(s, a)\right] \nonumber \\
&+&\kappa \mathbb{E}_{s \sim \rho_{\mathcal D}(\cdot), a \sim \pi' (\cdot \mid s)}\left[A_{\pi_k}(s, a)\right] \nonumber \\
&=&\kappa \mathbb{E}_{s \sim \rho_{\mathcal D}(\cdot), a \sim \pi'(\cdot|s)}[A_{\pi_k}(s,a)]
\end{eqnarray}
\end{theorem}

Theorem \ref{thm:bppo_for_ref} establishes that the performance difference $J_{\Delta}\left(\pi, \pi_k\right)$ between two consecutive reference policies can be lower bounded by its offline approximation $\widehat{J}_{\Delta}\left(\pi, \pi_k\right)$ and some total variance (TV) divergence terms. The first of the TV terms represents the discrepancy between the new learned policy $\pi'$ and the old reference policy $\pi_k$, the second TV term represents the distribution shift of the old reference policy $\pi_k$ from the approximated behavior policy $\pi_b$, and the third TV term can be seen as the approximation error of the behavior policy $\pi_b$. Note that only the first of the TV terms is related to $\pi'$, while the other two TV terms can be seen as constants. Thus, we have the following conclusion.

\begin{corollary}\label{thm:conclusion}
To ensure monotonic improvement in the expected discounted return under the offline setting, it suffices to maintain positive and maximize the lower bound in Eq.~\eqref{eq:DRImprovementLowerBound}, which corresponds to simultaneously maximizing $\mathbb{E}_{s \sim {\mathcal D}, a \sim \pi'(\cdot|s)}\left[A_{\pi_k}\left(s,a\right)\right]$ and minimizing $\max_{s} \left[ D_{TV}\left(\pi' \| \pi_k\right)\left[s\right] \right]$.
\end{corollary}

\begin{table*}[ht]
\centering
\setlength{\tabcolsep}{2.2pt}
\caption{Comparison of the performance of different algorithms. We report the average normalized score of 10 trajectories collected by the final learned policy on 5 seeds. HC=Halfcheetah, Hop=Hopper, W=Walker, m=medium-v2, mr=medium-replay-v2, me=medium-expert-v2, e=expert-v2.
}
\label{tab:d4rl-mujoco}
\begin{tabular}{c|cccc|cc|ccc}
\toprule
\multirow{2}{*}{Dataset} & \multicolumn{4}{c|}{Policy Constraint} & \multicolumn{2}{c|}{Value Conservative} & \multicolumn{3}{c}{Proposed} \\
\cmidrule{2-5} \cmidrule{6-7} \cmidrule{8-10}
& TD3+BC & ReBRAC & wPC & CPI-RE & SAC-RND & EDAC & TD3+SelfBC & TD3+SelfBC (select) & TD3+ESBC \\
\midrule
HC-m & 48.3 $\pm$ 0.3 & 65.6 $\pm$ 1.0 & 53.3 & 65.9 $\pm$ 1.6 & \textbf{66.6} $\pm$ 1.6 & 65.9 $\pm$ 0.6 & 61.0 $\pm$ 1.1 & 63.3 $\pm$ 0.2 & 63.5 $\pm$ 0.4 \\
HC-mr & 44.6 $\pm$ 0.5 & 51.0 $\pm$ 0.8 & 48.3 & 55.9 $\pm$ 1.5 & 54.9 $\pm$ 0.6 & \textbf{61.3} $\pm$ 1.9 & 52.3 $\pm$ 0.5 & 52.3 $\pm$ 0.6 & 54.8 $\pm$ 0.3 \\
HC-me & 90.7 $\pm$ 4.3 & 101.1 $\pm$ 5.2 & 93.7 & 95.6 $\pm$ 0.9 & \textbf{107.6} $\pm$ 2.8 & 106.3 $\pm$ 1.9 & 95.3 $\pm$ 3.1 & 96.5 $\pm$ 2.2 & 93.8 $\pm$ 2.9 \\
HC-e & 96.7 $\pm$ 1.1 & 105.9 $\pm$ 1.7 & - & 97.4 $\pm$ 0.4 & 105.8 $\pm$ 1.9 & \textbf{106.8} $\pm$ 3.4 & 100.9 $\pm$ 1.4 & 102.1 $\pm$ 0.5 & 102.0 $\pm$ 0.4 \\
\midrule
W-m & 83.7 $\pm$ 2.1 & 82.5 $\pm$ 3.6 & 86.0 & 86.3 $\pm$ 1.0 & 91.6 $\pm$ 2.8 & 92.5 $\pm$ 0.8 & 85.7 $\pm$ 17.9 & 99.1 $\pm$ 0.6 & \textbf{100.0} $\pm$ 1.4 \\
W-mr & 81.8 $\pm$ 5.5 & 77.3 $\pm$ 7.9 & 89.9 & 93.8 $\pm$ 2.2 & 88.7 $\pm$ 7.7 & 87.1 $\pm$ 2.3 & 95.4 $\pm$ 4.8 & 99.4 $\pm$ 1.4 & \textbf{100.4} $\pm$ 2.2 \\
W-me & 110.1 $\pm$ 0.5 & 111.6 $\pm$ 0.3 & 110.1 & 111.2 $\pm$ 0.5 & 105.0 $\pm$ 7.9 & \textbf{114.7} $\pm$ 0.9 & 112.0 $\pm$ 0.2 & 112.0 $\pm$ 0.2 & 112.9 $\pm$ 0.3 \\
W-e & 110.2 $\pm$ 0.3 & 112.3 $\pm$ 0.2 & - & 111.2 $\pm$ 0.2 & 114.3 $\pm$ 0.6 & \textbf{115.1} $\pm$ 1.9 & 112.6 $\pm$ 0.3 & 112.9 $\pm$ 0.1 & 113.1 $\pm$ 0.1 \\
\midrule
Hop-m & 59.3 $\pm$ 4.2 & 102.0 $\pm$ 1.0 & 86.5 & 97.9 $\pm$ 4.4 & 97.8 $\pm$ 2.3 & 101.6 $\pm$ 0.6 & 102.9 $\pm$ 0.1 & \textbf{103.2} $\pm$ 0.2 & 102.9 $\pm$ 0.1 \\
Hop-mr & 60.9 $\pm$ 18.8 & 98.1 $\pm$ 5.3 & 97.0 & 103.2 $\pm$ 1.4 & 100.5 $\pm$ 1.0 & 101.0 $\pm$ 0.5 & 101.7 $\pm$ 1.6 & \textbf{103.6} $\pm$ 0.4 & 102.6 $\pm$ 0.6 \\
Hop-me & 98.0 $\pm$ 9.4 & 107.0 $\pm$ 6.4 & 95.7 & 110.1 $\pm$ 4.1 & 109.8 $\pm$ 0.6 & 110.7 $\pm$ 0.1 & 97.2 $\pm$ 16.7 & \textbf{110.8} $\pm$ 2.6 & 99.1 $\pm$ 5.5 \\
Hop-e & 107.8 $\pm$ 7.0 & 100.1 $\pm$ 8.3 & - & 102.3 $\pm$ 0.5 & 109.7 $\pm$ 0.3 & 110.1 $\pm$ 0.1 & 112.4 $\pm$ 1.4 & \textbf{112.4} $\pm$ 0.4 & 111.1 $\pm$ 1.8 \\
\midrule
Avg on m and m-r & 63.1 & 79.4 & 76.8 & 83.8 & 83.3 & 84.9 & 83.2 & 86.8 & \textbf{87.4} \\
\midrule
Total avg & 82.7 & 92.9 & - & 94.2 & 96.0 & \textbf{97.8} & 94.1 & 97.3 & 96.4 \\
\bottomrule
\end{tabular}
\end{table*}

\paragraph{The connection between our TD3+SelfBC and Conclusion \ref{thm:conclusion}}
\begin{itemize}
\item Conservative policy update: the EMA process of TD3+SelfBC in Eq.~\eqref{eq:ref_EMA} can be seen as a conservative update step of the reference policy in Definition \ref{thm:def_pi}.
\item Policy learning: the TD3+SelfBC objective in Eq.~\eqref{eq:TD3+SelfBC} can be transformed to $\mathbb{E}_{s \in D}\left[\alpha Q_{\pi'}\left(s,\pi'\left(s\right)\right) - \left(\pi'\left(s\right)-\pi_k\left(s\right)\right)^2\right]$ using Definition \ref{thm:def_pi}, which learns a policy $\pi'$ in almost the same way as Conclusion \ref{thm:conclusion}. Thus, TD3+SelfBC can achieve a nearly monotonically improved reference policy.
\item The only non-rigorous step is that the Q term of TD3+SelfBC is $Q_{\pi'}$, but it should be $Q_{\pi_k}$ in theory. Nevertheless, as both the reference policy and the learned policy are updated mutually and dynamically, the difference between $Q_{\pi'}$ and $Q_{\pi_k}$ is small.
\end{itemize}

\vspace*{-0.2in}
\section{Experiments}
Our experiments are designed to evaluate the performance of the proposed algorithm against existing offline RL algorithms. We analyze the impact of the hyperparameter involved by our algorithm, namely the update ratio of the reference policy, on its performance. In addition, we conduct ablation studies to highlight the contribution of individual components within our algorithm to its overall performance.

\vspace*{-0.15in}
\subsection{Results on D4RL Benchmarks}
In Table \ref{tab:d4rl-mujoco}, we evaluate our algorithm in the D4RL \citep{fu2020d4rl} MuJoCo domain and compare with several previous policy constraint methods, including TD3+BC \citep{fujimoto2021td3bc}, ReBRAC \citep{tarasov2024rebrac}, wPC \citep{peng2023wPC}, CPI-RE \citep{hu2023cpi_re}, and we also compare with the value conservative methods SAC-RND \citep{nikulin2023sac-rnd} and EDAC \citep{an2021edac}. 
We provide additional details for these baselines in Appendix \ref{app:baselines}. 
We evaluate the performance of the learned policy by collecting 10 trajectories each, after 5000 training steps, and compute the average normalized score. We report the mean and standard derivation of the final learned policy's performance on five seeds (50 trajectories in total). The results of TD3+BC, ReBRAC, wPC, CPI-RE, SAC-RND, and EDAC are all obtained from their corresponding original publications mentioned earlier. 

Extensive experiments show that our algorithm achieves state-of-the-art performance among the class of policy constraint methods for offline RL. In particular, our algorithm demonstrates significant performance gains on the datasets containing non-expert trajectories, such as the medium and medium-replay datasets, highlighting the effectiveness of our proposed self-constraint mechanism. For the datasets containing expert trajectories, although there is little room for improvement, our algorithm can still make slight improvements.

For TD3+SelfBC, we report the evaluations on two settings:
\begin{itemize}
\vspace{-0.6em}
    \item TD3+SelfBC: we evaluate our algorithm on 5 different seeds, the pre-trained results loaded by each run are trained on different seeds (5 in total).
    \item TD3+SelfBC (select):  we also evaluate our algorithm on 5 different seeds, however, the pre-trained results loaded by all the 5 runs are the same, which are manually selected for best performance.
\vspace{-0.6em}
\end{itemize}

To obtain pre-trained TD3+EBC results, we set $N_{\textnormal{EBC}}=1e6$ for the expert and medium-expert datasets of halfcheetah and hopper, $N_{\textnormal{EBC}}=2e5$ for other datasets; we set $\alpha=1.0$ for all expert and medium-expert datasets and $\alpha=2.5$ for all the medium and medium-replay datasets, and the settings of $\alpha$ are also used by TD3+SelfBC and TD3+ESBC. Most of the hyperparameters of our algorithms follow TD3+BC. Full details of the hyperparameters are provided in Appendix \ref{app:hyperparameters}.

For TD3+ESBC, we load $N_{\textnormal{ens}}=5$ pre-trained TD3+EBC results that were trained on different seeds.
We only evaluate the first one of the ensemble policies simultaneously learned by TD3+ESBC. 

\vspace*{-0.1in}
\subsection{Analysis of the Reference Policy Update Ratio}

Using the TD3+SelfBC (select) experiments, we study how $\tau_{\textnormal{ref}}$, the update ratio of the reference policy, affects the performance of the proposed algorithm.

\begin{figure*}[t]
    \centering
    \begin{subfigure}{.45\linewidth}
        \setlength{\abovecaptionskip}{5pt}
        \setlength{\belowcaptionskip}{10pt}
        \includegraphics[width=1.0\linewidth]{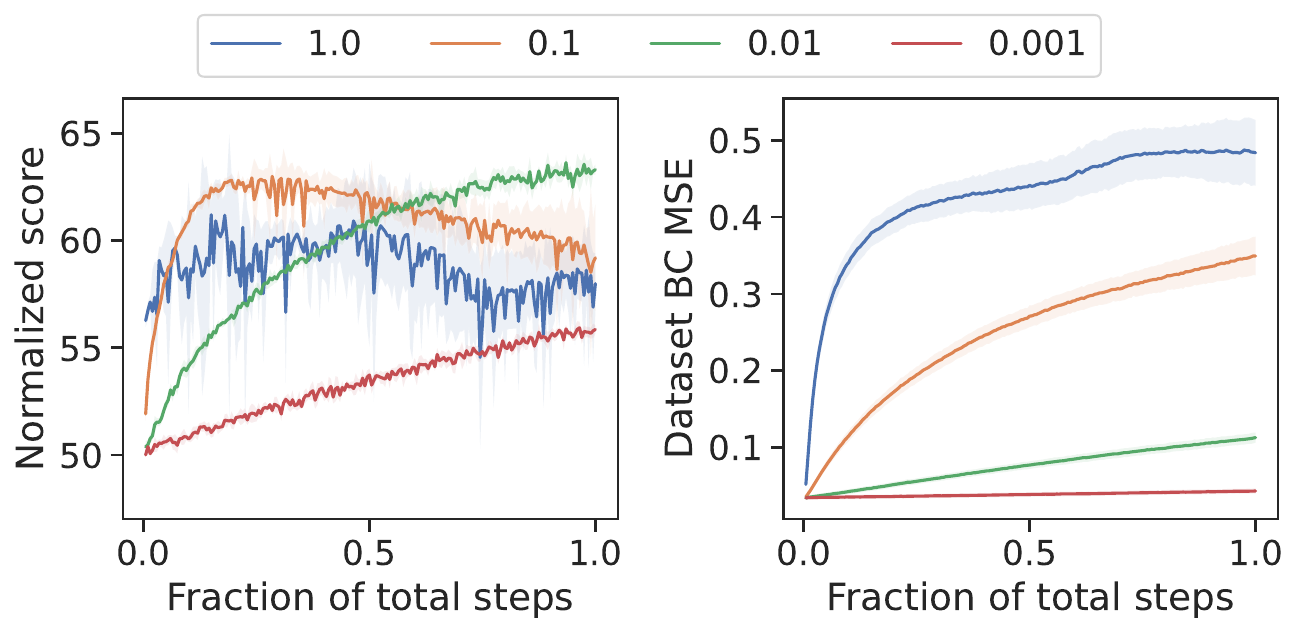}
        \caption{halfcheetah-medium-v2}
    \end{subfigure}
    \begin{subfigure}{.45\linewidth}
        \setlength{\abovecaptionskip}{5pt}
        \setlength{\belowcaptionskip}{10pt}
        \includegraphics[width=1.0\linewidth]{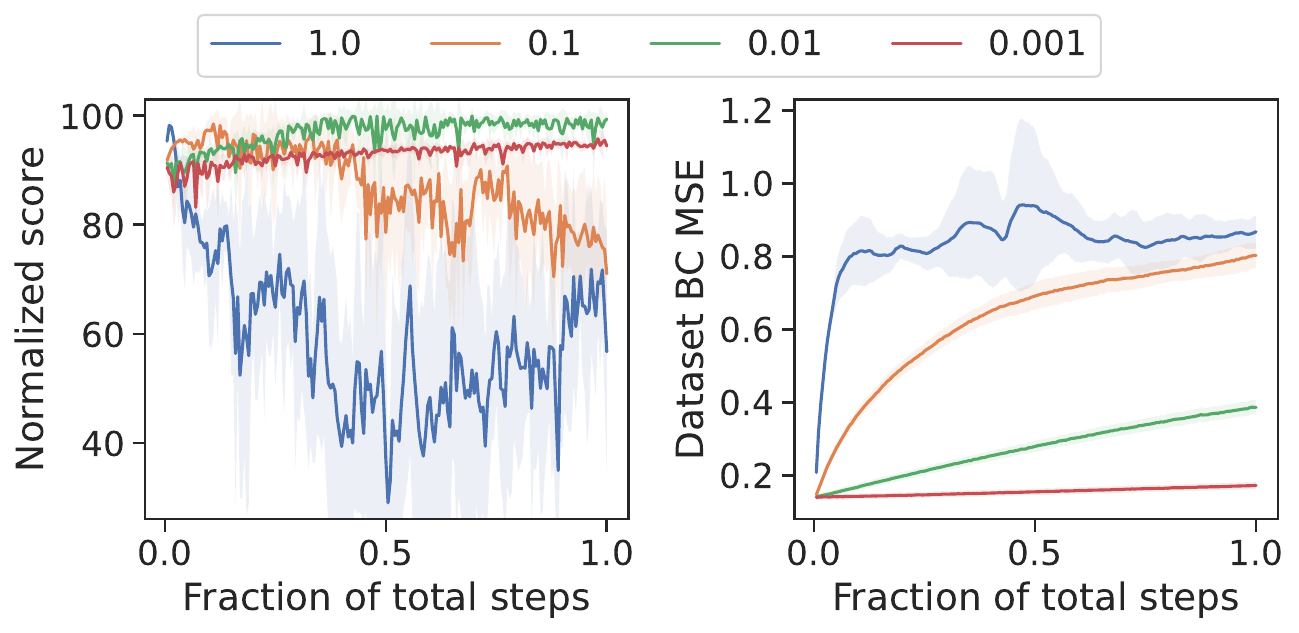}
        \caption{walker2d-medium-replay-v2}
    \end{subfigure}
    \setlength{\belowcaptionskip}{8pt}
    \caption{The effect of $scale_{\textnormal{ref}}$ on performance. We show the normalized scores and the dataset BC MSEs during the training process of TD3+SelfBC.}
    \label{fig:ablation_tau}
\end{figure*}

We compute $\tau_{\textnormal{ref}}$ by the following equation $\tau_{\textnormal{ref}}=\tau * {scale}_{\textnormal{ref}}$, where $\tau$ is the update ratio of the target policy introduced by TD3+BC, and we adjust the ${scale}_{\textnormal{ref}}$ in [1.0, 0.1, 0.01, 0.001]. In Fig. \ref{fig:ablation_tau}, we report the curves of the normalized score and the dataset BC MSE (distribution shift between the learned policy and the offline dataset) during the training process of TD3+SelfBC. We observe that a larger ${scale}_{\textnormal{ref}}$ results in a faster growth rate of the distribution shift. However, when the distribution shift is too large, the normalized score will decrease. In addition, a too small ${scale}_{\textnormal{ref}}$ will slow down the learning process. Thus, ${scale}_{\textnormal{ref}}$ should be set properly to balance the learning speed and the distribution shift.

\begin{figure*}[ht]
    \centering
    \setlength{\abovecaptionskip}{5pt}
    \setlength{\belowcaptionskip}{8pt}
    \includegraphics[width=0.95\linewidth]{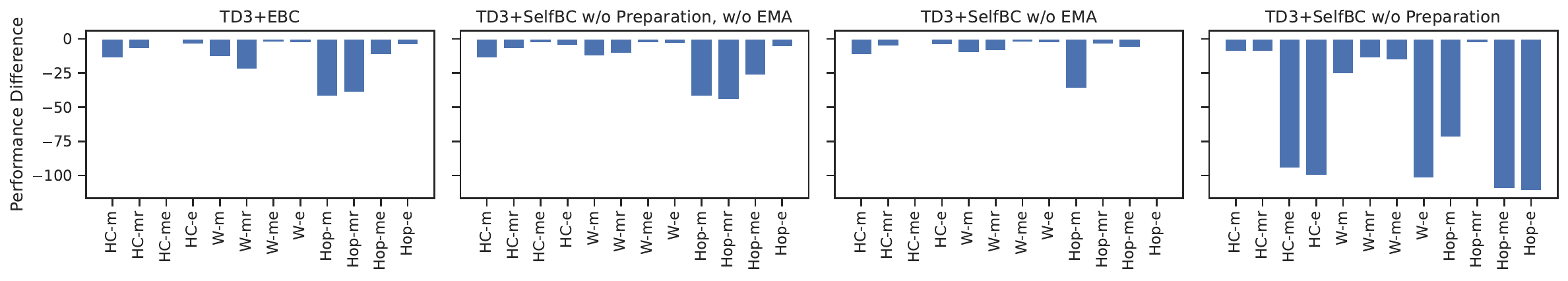}
    \caption{Performance difference of TD3+EBC and three variants of TD3+SelfBC compared with TD3+SelfBC.}
    \label{fig:ablation-bar-prepare_self}
\end{figure*}

\begin{figure}
    \centering
    \setlength{\abovecaptionskip}{5pt}
    \setlength{\belowcaptionskip}{8pt}
    \includegraphics[width=0.7\linewidth]{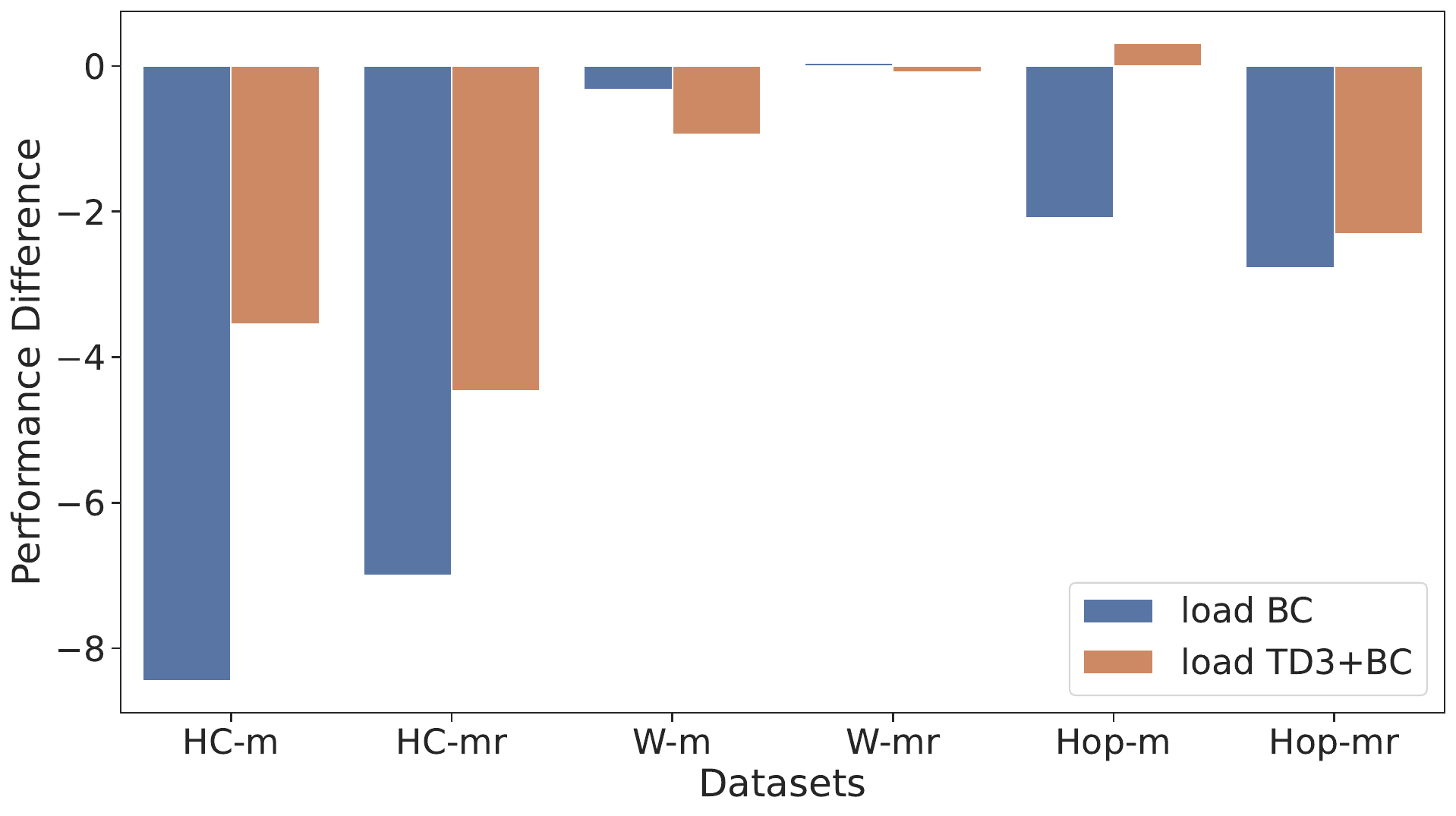}
    \caption{Performance difference of TD3+SelfBC variants with alternative pretrain algorithms compared with the original TD3+SelfBC that uses TD3+EBC.}
    \label{fig:appendix-seaborn_bar-ablation_load}
\end{figure}

\vspace*{-0.1in}
\subsection{Ablation Studies}\label{sec:ablation}

Using the TD3+SelfBC (select) experiments, we conducted a number of ablation studies to  highlight the contribution of individual components within our algorithm to its overall performance.

Our algorithm in Algorithm~\ref{alg:TD3+SelfBC} can be divided into three phases: loading the pre-trained results, TD3+SelfBC training preparation and TD3+SelfBC training. We remove or change one of the three phases to obtain the variants, and all the variants are as follows:

\textbf{TD3+SelfBC w/o Preparation.} In this variant, we modify the preparation phase by updating the reference policy network $\pi_{\tilde{\theta}}$ to be the behavior policy $\pi_b$ rather than the pre-trained policy.

\textbf{TD3+SelfBC w/o EMA.} In this variant, the reference policy $\pi_{\tilde{\theta}}$ in the TD3+SelfBC training phase is fixed instead of updated by EMA.

\textbf{TD3+SelfBC w/o Preparation, w/o EMA.} This variant combines the two variants 
mentioned above.

\textbf{Pretraining} We initialize the training of TD3+SelfBC by loading the pre-trained results from previous policy constraint methods, which can be BC, TD3+BC and TD3+EBC. 
To analyze the effect of pretraining algorithms, we introduce the following two variants: load BC and load TD3+BC. 
For TD3+BC, we run the code from CORL \citep{tarasov2022corl} without any modification. For BC, we first train a behavior policy by supervised learning and then train the Q function of the behavior policy; we implement it by training a variant of TD3+BC without policy loss and fix the learned policy to be the behavior policy. For TD3+EBC, we implement the version in Algorithm~\ref{alg:TD3+SelfBC}.

In Fig. \ref{fig:ablation-bar-prepare_self}, we compare TD3+SelfBC with its three variants and TD3+EBC and show the performance difference. TD3+SelfBC w/o Preparation has the biggest gap, because EMA requires the parameters of the source model and target model to be similar, otherwise the parameter-update will be unstable, which leads to an OOD policy. TD3+SelfBC w/o Preparation w/o EMA performs similar with TD3+EBC, as the biggest difference between them is the total training epochs. TD3+SelfBC w/o EMA performs better than TD3+EBC, especially for non-expert datasets, where it is in line with our expectations because the preparation phase contains a hard-update step of $\pi_{\tilde{\theta}}$.
Because there is little improvement space for the datasets that contain expert trajectories, TD3+SelfBC w/o EMA performs similar to TD3+SelfBC on the medium-expert and expert datasets.

In Fig. \ref{fig:appendix-seaborn_bar-ablation_load}, we report the performance difference of the two variants compared with the original TD3+SelfBC, which leverages the pre-trained results of TD3+EBC.

\section{Conclusion}
In this work, we empirically demonstrate the limitation of TD3+BC and attribute it to the reliance on the static reference policy. We summarized previously weighted behavior cloning as a special design of the reference policy for the general behavior cloning. To address these limitations, we propose Self Behavior Cloning (SelfBC), a novel dynamic policy constraint for offline reinforcement learning. SelfBC progressively updates the reference policy towards the learned policy during training. By integrating this dynamic self-constraint mechanism into the off-policy algorithm TD3, our TD3+SelfBC method can learn an non-conservative policy and avoid the policy collapse. We theoretically show that our conservatively updated reference policy can achieve nearly monotonic improvement. Extensive experiments on the D4RL MuJoCo benchmark have demonstrated the excellent performance of our approach. In the future, we will explore alternative mechanisms for updating the reference policy and extend SelfBC to learn stochastic policies.


\begin{ack}
This work is supported by the National Natural Science Foundation of China (No. 62071154, 62306242).
\end{ack}

\bibliography{mybibfile}

\begin{thebibliography}{40}
\providecommand{\natexlab}[1]{#1}
\providecommand{\url}[1]{\texttt{#1}}
\expandafter\ifx\csname urlstyle\endcsname\relax
  \providecommand{\doi}[1]{doi: #1}\else
  \providecommand{\doi}{doi: \begingroup \urlstyle{rm}\Url}\fi

\bibitem[Achiam et~al.(2017)Achiam, Held, Tamar, and Abbeel]{achiam2017constrainedpo}
J.~Achiam, D.~Held, A.~Tamar, and P.~Abbeel.
\newblock Constrained policy optimization.
\newblock In \emph{International conference on machine learning}, pages 22--31. PMLR, 2017.

\bibitem[An et~al.(2021)An, Moon, Kim, and Song]{an2021edac}
G.~An, S.~Moon, J.-H. Kim, and H.~O. Song.
\newblock Uncertainty-based offline reinforcement learning with diversified {Q}-ensemble.
\newblock \emph{Advances in neural information processing systems}, 34:\penalty0 7436--7447, 2021.

\bibitem[Bai et~al.(2022)Bai, Wang, Yang, Deng, Garg, Liu, and Wang]{bai2022pbrl}
C.~Bai, L.~Wang, Z.~Yang, Z.~Deng, A.~Garg, P.~Liu, and Z.~Wang.
\newblock Pessimistic bootstrapping for uncertainty-driven offline reinforcement learning.
\newblock In \emph{10th International Conference on Learning Representations, ICLR 2022}, 2022.

\bibitem[Beeson and Montana(2022)]{beeson2022improvingtd3bc}
A.~Beeson and G.~Montana.
\newblock Improving {TD3-BC}: Relaxed policy constraint for offline learning and stable online fine-tuning.
\newblock In \emph{3rd Offline RL Workshop: Offline RL as a''Launchpad''}, 2022.

\bibitem[Caron et~al.(2021)Caron, Touvron, Misra, J{\'e}gou, Mairal, Bojanowski, and Joulin]{caron2021dino}
M.~Caron, H.~Touvron, I.~Misra, H.~J{\'e}gou, J.~Mairal, P.~Bojanowski, and A.~Joulin.
\newblock Emerging properties in self-supervised vision transformers.
\newblock In \emph{Proceedings of the IEEE/CVF international conference on computer vision}, pages 9650--9660, 2021.

\bibitem[Chen et~al.(2020)Chen, Zhou, Wang, Wang, Wu, and Ross]{chen2020bail}
X.~Chen, Z.~Zhou, Z.~Wang, C.~Wang, Y.~Wu, and K.~Ross.
\newblock {BAIL}: Best-action imitation learning for batch deep reinforcement learning.
\newblock \emph{Advances in Neural Information Processing Systems}, 33:\penalty0 18353--18363, 2020.

\bibitem[El~Sallab et~al.(2017)El~Sallab, Abdou, Perot, and Yogamani]{sallab2017autodriving}
A.~El~Sallab, M.~Abdou, E.~Perot, and S.~Yogamani.
\newblock Deep reinforcement learning framework for autonomous driv-ing.
\newblock \emph{stat}, 1050:\penalty0 8, 2017.

\bibitem[Fakoor et~al.(2021)Fakoor, Mueller, Asadi, Chaudhari, and Smola]{fakoor2021CDC_sac+bc}
R.~Fakoor, J.~W. Mueller, K.~Asadi, P.~Chaudhari, and A.~J. Smola.
\newblock Continuous doubly constrained batch reinforcement learning.
\newblock \emph{Advances in Neural Information Processing Systems}, 34:\penalty0 11260--11273, 2021.

\bibitem[Fu et~al.(2020)Fu, Kumar, Nachum, Tucker, and Levine]{fu2020d4rl}
J.~Fu, A.~Kumar, O.~Nachum, G.~Tucker, and S.~Levine.
\newblock {D4RL}: Datasets for deep data-driven reinforcement learning.
\newblock \emph{arXiv preprint arXiv:2004.07219}, 2020.

\bibitem[Fujimoto and Gu(2021)]{fujimoto2021td3bc}
S.~Fujimoto and S.~S. Gu.
\newblock A minimalist approach to offline reinforcement learning.
\newblock \emph{Advances in neural information processing systems}, 34:\penalty0 20132--20145, 2021.

\bibitem[Fujimoto et~al.(2018)Fujimoto, Hoof, and Meger]{fujimoto2018td3}
S.~Fujimoto, H.~Hoof, and D.~Meger.
\newblock Addressing function approximation error in actor-critic methods.
\newblock In \emph{International conference on machine learning}, pages 1587--1596. PMLR, 2018.

\bibitem[He et~al.(2020)He, Fan, Wu, Xie, and Girshick]{he2020moco}
K.~He, H.~Fan, Y.~Wu, S.~Xie, and R.~Girshick.
\newblock Momentum contrast for unsupervised visual representation learning.
\newblock In \emph{Proceedings of the IEEE/CVF conference on computer vision and pattern recognition}, pages 9729--9738, 2020.

\bibitem[Hu et~al.(2024)Hu, Ma, Xiao, Zheng, and Hao]{hu2023cpi_re}
X.~Hu, Y.~Ma, C.~Xiao, Y.~Zheng, and J.~Hao.
\newblock Iteratively refined behavior regularization for offline reinforcement learning.
\newblock In \emph{NeurIPS 2023 Workshop on Distribution Shifts: New Frontiers with Foundation Models}, 2024.

\bibitem[Kakade and Langford(2002)]{kakade2002cpi}
S.~Kakade and J.~Langford.
\newblock Approximately optimal approximate reinforcement learning.
\newblock In \emph{Proceedings of the Nineteenth International Conference on Machine Learning}, pages 267--274, 2002.

\bibitem[Kidambi et~al.(2020)Kidambi, Rajeswaran, Netrapalli, and Joachims]{kidambi2020morel}
R.~Kidambi, A.~Rajeswaran, P.~Netrapalli, and T.~Joachims.
\newblock {MoRel}: Model-based offline reinforcement learning.
\newblock \emph{Advances in neural information processing systems}, 33:\penalty0 21810--21823, 2020.

\bibitem[Kostrikov et~al.(2021)Kostrikov, Fergus, Tompson, and Nachum]{kostrikov2021fisher-brc}
I.~Kostrikov, R.~Fergus, J.~Tompson, and O.~Nachum.
\newblock Offline reinforcement learning with {F}isher divergence critic regularization.
\newblock In \emph{International Conference on Machine Learning}, pages 5774--5783. PMLR, 2021.

\bibitem[Kumar et~al.(2019)Kumar, Fu, Soh, Tucker, and Levine]{kumar2019BEAR_MMD}
A.~Kumar, J.~Fu, M.~Soh, G.~Tucker, and S.~Levine.
\newblock Stabilizing off-policy {Q}-learning via bootstrapping error reduction.
\newblock \emph{Advances in neural information processing systems}, 32, 2019.

\bibitem[Levine et~al.(2020)Levine, Kumar, Tucker, and Fu]{levine2020offline}
S.~Levine, A.~Kumar, G.~Tucker, and J.~Fu.
\newblock Offline reinforcement learning: Tutorial, review, and perspectives on open problems.
\newblock \emph{arXiv preprint arXiv:2005.01643}, 2020.

\bibitem[Mnih et~al.(2015)Mnih, Kavukcuoglu, Silver, Rusu, Veness, Bellemare, Graves, Riedmiller, Fidjeland, Ostrovski, et~al.]{mnih2015dqn}
V.~Mnih, K.~Kavukcuoglu, D.~Silver, A.~A. Rusu, J.~Veness, M.~G. Bellemare, A.~Graves, M.~Riedmiller, A.~K. Fidjeland, G.~Ostrovski, et~al.
\newblock Human-level control through deep reinforcement learning.
\newblock \emph{nature}, 518\penalty0 (7540):\penalty0 529--533, 2015.

\bibitem[Morales-Brotons et~al.(2024)Morales-Brotons, Vogels, and Hendrikx]{morales-brotons2024ema_in_dl}
D.~Morales-Brotons, T.~Vogels, and H.~Hendrikx.
\newblock Exponential moving average of weights in deep learning: Dynamics and benefits.
\newblock \emph{Transactions on Machine Learning Research}, 2024.
\newblock ISSN 2835-8856.

\bibitem[Nair et~al.(2020)Nair, Dalal, Gupta, and Levine]{ashvin2020awac}
A.~Nair, M.~Dalal, A.~Gupta, and S.~Levine.
\newblock Accelerating online reinforcement learning with offline datasets.
\newblock \emph{CoRR, vol. abs/2006.09359}, 2020.

\bibitem[Nikulin et~al.(2023)Nikulin, Kurenkov, Tarasov, and Kolesnikov]{nikulin2023sac-rnd}
A.~Nikulin, V.~Kurenkov, D.~Tarasov, and S.~Kolesnikov.
\newblock Anti-exploration by random network distillation.
\newblock In \emph{International Conference on Machine Learning}, pages 26228--26244. PMLR, 2023.

\bibitem[Peng et~al.(2019)Peng, Kumar, Zhang, and Levine]{peng2019awr}
X.~B. Peng, A.~Kumar, G.~Zhang, and S.~Levine.
\newblock Advantage-weighted regression: Simple and scalable off-policy reinforcement learning.
\newblock \emph{arXiv preprint arXiv:1910.00177}, 2019.

\bibitem[Peng et~al.(2023)Peng, Han, Liu, and Zhou]{peng2023wPC}
Z.~Peng, C.~Han, Y.~Liu, and Z.~Zhou.
\newblock Weighted policy constraints for offline reinforcement learning.
\newblock In \emph{Proceedings of the AAAI Conference on Artificial Intelligence}, volume~37, pages 9435--9443, 2023.

\bibitem[Schulman et~al.(2015)Schulman, Levine, Abbeel, Jordan, and Moritz]{schulman2015trpo}
J.~Schulman, S.~Levine, P.~Abbeel, M.~Jordan, and P.~Moritz.
\newblock Trust region policy optimization.
\newblock In \emph{International conference on machine learning}, pages 1889--1897. PMLR, 2015.

\bibitem[Schulman et~al.(2017)Schulman, Wolski, Dhariwal, Radford, and Klimov]{schulman2017ppo}
J.~Schulman, F.~Wolski, P.~Dhariwal, A.~Radford, and O.~Klimov.
\newblock Proximal policy optimization algorithms.
\newblock \emph{arXiv preprint arXiv:1707.06347}, 2017.

\bibitem[Siegel et~al.(2020)Siegel, Springenberg, Berkenkamp, Abdolmaleki, Neunert, Lampe, Hafner, Heess, and Riedmiller]{siegel2020abm}
N.~Siegel, J.~T. Springenberg, F.~Berkenkamp, A.~Abdolmaleki, M.~Neunert, T.~Lampe, R.~Hafner, N.~Heess, and M.~Riedmiller.
\newblock Keep doing what worked: Behavior modelling priors for offline reinforcement learning.
\newblock In \emph{International Conference on Learning Representations}, 2020.

\bibitem[Sutton and Barto(2018)]{sutton2018rlintroduction}
R.~S. Sutton and A.~G. Barto.
\newblock \emph{Reinforcement learning: An introduction}.
\newblock MIT press, 2018.

\bibitem[Tarasov et~al.(2023)Tarasov, Nikulin, Akimov, Kurenkov, and Kolesnikov]{tarasov2022corl}
D.~Tarasov, A.~Nikulin, D.~Akimov, V.~Kurenkov, and S.~Kolesnikov.
\newblock Corl: Research-oriented deep offline reinforcement learning library.
\newblock In A.~Oh, T.~Neumann, A.~Globerson, K.~Saenko, M.~Hardt, and S.~Levine, editors, \emph{Advances in Neural Information Processing Systems}, volume~36, pages 30997--31020. Curran Associates, Inc., 2023.

\bibitem[Tarasov et~al.(2024)Tarasov, Kurenkov, Nikulin, and Kolesnikov]{tarasov2024rebrac}
D.~Tarasov, V.~Kurenkov, A.~Nikulin, and S.~Kolesnikov.
\newblock Revisiting the minimalist approach to offline reinforcement learning.
\newblock \emph{Advances in Neural Information Processing Systems}, 36, 2024.

\bibitem[Vieillard et~al.(2020)Vieillard, Pietquin, and Geist]{vieillard2020deepcpi}
N.~Vieillard, O.~Pietquin, and M.~Geist.
\newblock Deep conservative policy iteration.
\newblock In \emph{Proceedings of the AAAI Conference on Artificial Intelligence}, volume~34, pages 6070--6077, 2020.

\bibitem[Wang et~al.(2018)Wang, Xiong, Han, Liu, Zhang, et~al.]{wang2018marwil}
Q.~Wang, J.~Xiong, L.~Han, H.~Liu, T.~Zhang, et~al.
\newblock Exponentially weighted imitation learning for batched historical data.
\newblock \emph{Advances in Neural Information Processing Systems}, 31, 2018.

\bibitem[Wang et~al.(2020)Wang, Novikov, Zolna, Merel, Springenberg, Reed, Shahriari, Siegel, Gulcehre, Heess, et~al.]{wang2020crr}
Z.~Wang, A.~Novikov, K.~Zolna, J.~S. Merel, J.~T. Springenberg, S.~E. Reed, B.~Shahriari, N.~Siegel, C.~Gulcehre, N.~Heess, et~al.
\newblock Critic regularized regression.
\newblock \emph{Advances in Neural Information Processing Systems}, 33:\penalty0 7768--7778, 2020.

\bibitem[Wu et~al.(2022)Wu, Wu, Qiu, Wang, and Long]{wu2022spot}
J.~Wu, H.~Wu, Z.~Qiu, J.~Wang, and M.~Long.
\newblock Supported policy optimization for offline reinforcement learning.
\newblock \emph{Advances in Neural Information Processing Systems}, 35:\penalty0 31278--31291, 2022.

\bibitem[Wu et~al.(2021)Wu, Zhai, Srivastava, Susskind, Zhang, Salakhutdinov, and Goh]{wu2021uwac}
Y.~Wu, S.~Zhai, N.~Srivastava, J.~M. Susskind, J.~Zhang, R.~Salakhutdinov, and H.~Goh.
\newblock Uncertainty weighted actor-critic for offline reinforcement learning.
\newblock In \emph{International Conference on Machine Learning}, pages 11319--11328. PMLR, 2021.

\bibitem[Yu et~al.(2021{\natexlab{a}})Yu, Liu, Nemati, and Yin]{yu2021healthcare}
C.~Yu, J.~Liu, S.~Nemati, and G.~Yin.
\newblock Reinforcement learning in healthcare: A survey.
\newblock \emph{ACM Computing Surveys (CSUR)}, 55\penalty0 (1):\penalty0 1--36, 2021{\natexlab{a}}.

\bibitem[Yu et~al.(2020)Yu, Thomas, Yu, Ermon, Zou, Levine, Finn, and Ma]{yu2020mopo}
T.~Yu, G.~Thomas, L.~Yu, S.~Ermon, J.~Y. Zou, S.~Levine, C.~Finn, and T.~Ma.
\newblock {MOPO}: Model-based offline policy optimization.
\newblock \emph{Advances in Neural Information Processing Systems}, 33:\penalty0 14129--14142, 2020.

\bibitem[Yu et~al.(2021{\natexlab{b}})Yu, Kumar, Rafailov, Rajeswaran, Levine, and Finn]{yu2021combo}
T.~Yu, A.~Kumar, R.~Rafailov, A.~Rajeswaran, S.~Levine, and C.~Finn.
\newblock {COMBO}: Conservative offline model-based policy optimization.
\newblock \emph{Advances in neural information processing systems}, 34:\penalty0 28954--28967, 2021{\natexlab{b}}.

\bibitem[Zhang et~al.(2021)Zhang, Kuppannagari, and Viktor]{zhang2021brac+}
C.~Zhang, S.~Kuppannagari, and P.~Viktor.
\newblock {BRAC+}: Improved behavior regularized actor critic for offline reinforcement learning.
\newblock In \emph{Asian Conference on Machine Learning}, pages 204--219. PMLR, 2021.

\bibitem[Zhuang et~al.(2023)Zhuang, LEI, Liu, Wang, and Guo]{zhuang2023behaviorppo}
Z.~Zhuang, K.~LEI, J.~Liu, D.~Wang, and Y.~Guo.
\newblock Behavior proximal policy optimization.
\newblock In \emph{The Eleventh International Conference on Learning Representations}, 2023.

\end{thebibliography}

\newpage

\appendix

\section{Proof of the equivalence between Eqs.~\eqref{eq:wbc} and~\eqref{eq:general_wbc}}\label{appendix:equivalent_wbc}
In this section, we prove the equivalence between Eq.~\eqref{eq:wbc} and Eq.~\eqref{eq:general_wbc} by demonstrating that they have the equivalent gradient in expectation with respect to the learned policy $\pi_\theta$. It should be noted that we consider the deterministic $\pi_\theta$, but we do not restrict the ground truth behavior policy $\pi_b^{\mathcal{D}}$, which can be deterministic or stochastic, a single policy or mixed policies.

\begin{proof}
We first rewrite Eq.~\eqref{eq:wbc} as an integral formulation.
\begin{eqnarray}\label{eq:wbc_integral}
\mathbb{E}_{s \sim \mathcal{D}} \left[ \int \left( \pi_b^{\mathcal{D}}\left(a|s\right) \cdot w(s,a) \right) \cdot \left(\pi_\theta\left(s\right) - a\right)^2 da \right]
\end{eqnarray}
Then we define
\begin{eqnarray}\label{eq:qw}
q_{w}\left(a|s\right) = \frac{\pi_b^{\mathcal{D}}\left(a|s\right) w(s,a)}{C}
\end{eqnarray}
where $C$ is a normalization constant, which can be calculated as
\begin{eqnarray}\label{eq:c_definition}
C = \int \pi_b^{\mathcal{D}}\left(a|s\right) w(s,a) da
\end{eqnarray}
We substitute Eq.~\eqref{eq:qw} into Eq.~\eqref{eq:wbc_integral} and obtain:
\begin{eqnarray}
\mathbb{E}_{s \sim \mathcal{D}} \left[ \int \left(C \cdot q_w(a|s)\right) \cdot \left(\pi_\theta\left(s\right) - a\right)^2 da \right]
\end{eqnarray}
Rewriting this integral in its expectation formulation, we have
\begin{eqnarray}\label{eq:wbc_equal}
C \cdot \mathbb{E}_{s \sim \mathcal{D}, a \sim q_w\left(a|s\right)} \left[ \left(\pi_\theta\left(s\right) - a\right)^2 \right]
\end{eqnarray}

Now, we only need to prove that Eq.~\eqref{eq:wbc_equal} is equivalent to Eq.~\eqref{eq:general_wbc}. 
For simplicity, we consider a single state $s$ and ignore the normalization constant, the Eq.~\eqref{eq:wbc_equal} and Eq.~\eqref{eq:general_wbc} will become the terms $\rm A$ and $\rm B$, defined as:
\begin{eqnarray}
{\mathrm{A}}&=&\mathbb{E}_{a \sim q_w\left(s,a\right)} \left[ \left(\pi_\theta\left(s\right) - a\right)^2 \right] \\
\rm{B}&=&\left(\pi_\theta\left(s\right) - \mathbb{E}_{a \sim q_w\left(\cdot|s\right)}\left[a\right]\right)^2
\end{eqnarray}

Our objective now is to demonstrate the equivalence between $\rm A$ and $\rm B$. We then derive the connection between A and B:
\begin{eqnarray}\label{eq:connect_A_B}
{\mathrm{B}}&=&\left(\pi_\theta\left(s\right) - \mathbb{E}_{a \sim q_w\left(\cdot|s\right)}\left[a\right]\right)^2 \nonumber\\
&=&\left(\mathbb{E}_{a \sim q_w\left(\cdot|s\right)}\left[\pi_\theta\left(s\right) - a\right]\right)^2 \nonumber\\
&=&\mathbb{E}_{a \sim q_w\left(\cdot|s\right)}\left[\left(\pi_\theta\left(s\right) - a\right)^2\right] - {\mathrm{Var}}_{a \sim q_w\left(\cdot|s\right)}\left[\pi_\theta\left(s\right)-a\right] \nonumber\\
&=&{\mathrm{A}} - {\mathrm{Var}}_{a \sim q_w\left(\cdot|s\right)}\left[\pi_\theta\left(s\right)-a\right]
\end{eqnarray}

We notice that a new variance term ${\mathrm{Var}}_{a \sim q_w\left(\cdot|s\right)}\left[\pi_\theta\left(s\right)-a\right]$ can be seen in Eq.~\eqref{eq:connect_A_B}. 
Further analysis below shows that this variance term is independent of $\pi_\theta$:
\begin{eqnarray}\label{eq:var_derivation}
&&{\mathrm{Var}}_{a \sim q_w\left(\cdot|s\right)}\left[\pi_\theta\left(s\right)-a\right] \nonumber \\
=&&\mathbb{E}_{a \sim q_w\left(\cdot|s\right)}\left[\left(\left(\pi_\theta\left(s\right)-a\right)-\mathbb{E}_{a \sim q_w\left(\cdot|s\right)}\left[\pi_\theta\left(s\right)-a\right]\right)^2\right] \nonumber \\
=&&\mathbb{E}_{a \sim q_w\left(\cdot|s\right)}\left[\left(\pi_\theta\left(s\right)-a-\pi_\theta\left(s\right)+\mathbb{E}_{a \sim q_w\left(\cdot|s\right)}\left[a\right]\right)^2\right] \nonumber \\
=&&\mathbb{E}_{a \sim q_w\left(\cdot|s\right)}\left[\left(a-\mathbb{E}_{a \sim q_w\left(\cdot|s\right)}\left[a\right]\right)^2\right] \nonumber \\
=&&{\mathrm{Var}}_{a \sim q_w\left(\cdot|s\right)}\left[a\right]
\end{eqnarray}

The final expression in Eq.\eqref{eq:var_derivation} does not involve $\pi_\theta$, which represents the fact that the variance ${\mathrm{Var}}_{a \sim q_w(\cdot|s)}[a]$ is independent of $\pi_\theta$. By substituting Eq.\eqref{eq:var_derivation} back into Eq.\eqref{eq:connect_A_B}, we get:
\begin{eqnarray}
{\mathrm{B}}={\mathrm{A}} - {\mathrm{Var}}_{a \sim q_w\left(\cdot|s\right)}\left[a\right]
\end{eqnarray}
where the variance ${\mathrm{Var}}_{a \sim q_w\left(\cdot|s\right)}\left[a\right]$ is independent of $\pi_\theta$. 

Therefore, we have proved that $\mathrm{A}$ and $\mathrm{B}$ have the same gradient with respect to $\pi_\theta$:
\begin{eqnarray}\label{eq:grad_equal_A_B}
\nabla_\theta \mathrm{A} = \nabla_\theta \mathrm{B}
\end{eqnarray}

By adding back the expectation on $s$ in Eq.~\eqref{eq:grad_equal_A_B}, substituting Eq.~\eqref{eq:wbc} and Eq.~\eqref{eq:general_wbc} back into Eq.~\eqref{eq:grad_equal_A_B}, we have finally proved that:
\begin{eqnarray}\label{eq:wbc_grad_equivalence}
&& \frac{1}{C} \nabla_\theta \mathbb{E}_{s \sim \mathcal{D}, a \sim \pi_b^{\mathcal{D}}\left(\cdot|s\right)}[w(s,a) \cdot \left(\pi_\theta\left(s\right) - a\right)^2] \nonumber \\
&=& \nabla_\theta \mathbb{E}_{s \sim \mathcal{D}}\left[\left(\pi_\theta\left(s\right) - \mathbb{E}_{a \sim q_w\left(\cdot|s\right)}\left[a\right]\right)^2\right]
\end{eqnarray}
where $C$ is the normalization constant defined by Eq.~\eqref{eq:c_definition}.

\end{proof}

\section{Proof of the equivalence between EBC and BC}
\label{sec:appendix-proof_of_equivalence}
The equivalence between EBC and BC can be proved by assigning $w(s,a)=1$ to all $(s,a) \in \mathcal{D}$ for Eq.~\eqref{eq:wbc_grad_equivalence}. 
The constant $C$ and $q_w$ in Eq.~\eqref{eq:wbc_grad_equivalence} can be rewritten as $C=1$ and $q_w=\pi_b^{\mathcal{D}}$. 
Thus, Eq.~\eqref{eq:wbc_grad_equivalence} can be rewritten as:
\begin{eqnarray}\label{eq:ebc_grad_equivalence}
&& \nabla_\theta \mathbb{E}_{s \sim \mathcal{D}, a \sim \pi_b^{\mathcal{D}}(\cdot|s)}\left[\left(\pi_\theta\left(s\right) - a\right)^2\right] \nonumber \\
&=& \nabla_\theta \mathbb{E}_{s \sim \mathcal{D}}\left[\left(\pi_\theta\left(s\right) - \mathbb{E}_{a \sim \pi_b^{\mathcal{D}}\left(\cdot|s\right)}\left[a\right]\right)^2\right] \nonumber \\
&\approx& \nabla_\theta \mathbb{E}_{s \sim \mathcal{D}}\left[\left(\pi_\theta\left(s\right) - \pi_b\left(s\right)\right)^2\right]
\end{eqnarray}
where $\pi_b$ is the learned behavior policy using Eq.~\eqref{eq:pib}.

We demonstrate the equivalence between the EBC and the BC: when estimating the expectation-based objective using dataset samples, the gradient of the EBC objective $\nabla_\theta \hat{\mathcal{L}}_{\pi}^{\textnormal{EBC}}(\theta)$ represents the expectation of the gradient of the BC objective $\nabla_\theta \hat{\mathcal{L}}_{\pi}^{\textnormal{BC}}(\theta)$ with respect to dataset actions. Given a $(s,a)$ pair, we have the following sample-based gradient:
\begin{eqnarray}
 \mathbb{E}_{a \sim \pi_b^{\mathcal{D}}\left(\cdot|s\right)}\left[\nabla_\theta \hat{\mathcal{L}}_{\pi}^{\textnormal{BC}}\left(\theta\right)\left[s\right]\right] 
&\approx& \nabla_\theta \hat{\mathcal{L}}_{\pi}^{\textnormal{EBC}}(\theta)\left[s\right] \nonumber \\
\text{where } \hat{\mathcal{L}}_{\pi}^{\textnormal{BC}}(\theta)\left[s\right] &=& \left(\pi_\theta\left(s\right) - a\right)^2 \nonumber \\
\text{and } \hat{\mathcal{L}}_{\pi}^{\textnormal{EBC}}(\theta)\left[s\right] &=& \left(\pi_\theta\left(s\right) - \pi_b\left(s\right)\right)^2
\end{eqnarray}

Intuitively, EBC learns a policy to clone the behavior policy $\pi_b$, which is equivalent to cloning the dataset samples.

\begin{figure}[t]
\begin{algorithm}[H]
  \caption{TD3 + Expected Behavior Cloning (TD3+EBC)}
  \label{alg:seperateTD3+EBC}
\begin{algorithmic}
  \STATE {\bfseries Hyperparameters:} training steps $N_{\textnormal{BC}}, N_{\textnormal{EBC}}$, target update ratio $\tau$
  
  \STATE {\bfseries Initialize:} Q networks $Q_{\phi_1}, Q_{\phi_2}$ and target Q networks $Q_{\bar\phi_1} Q_{\bar \phi_2}$, policy network $\pi_\theta$ and target policy network $\pi_{\bar \theta}$, behavior policy network $\pi_b$
  
  \STATE // {\bfseries BC Training}
  \FOR{$t=1$ {\bfseries to} $N_{\textnormal{BC}}$}
    \STATE Sample mini-batch state-action pairs $\left(s, a\right)$ from $\mathcal{D}$
    \STATE Update $\pi_b$ by the objective in Eq.~\eqref{eq:pib}
  \ENDFOR
  
  \STATE // {\bfseries TD3+EBC Training}
  \FOR{$t=1$ {\bfseries to} $N_{\textnormal{EBC}}$}
    \STATE Sample mini-batch transitions $\left(s, a, r, s^{\prime}\right)$ from $\mathcal{D}$
    \STATE Update $\phi_{1,2}$ by minimizing $J_Q(\phi)$ in Eq.~\eqref{eq:Qloss_TD3+BC}
    \STATE Update $\theta$ by maximizing \\ \quad
    $J_{\pi}^{\textnormal{TD3+EBC}}(\theta)=\mathbb{E}_{s \sim \mathcal{D}}[\alpha Q(s, \pi_\theta(s)) - (\pi_\theta(s)-\pi_b(s))^2]$
    \STATE Update target networks: \\ \quad
    $\bar{\phi}_{1,2} \leftarrow \tau \phi_{1,2} + (1-\tau) \bar{\phi}_{1,2}$, $\bar{\theta} \leftarrow \tau \theta+(1-\tau) \bar{\theta}$
  \ENDFOR
\end{algorithmic}
\end{algorithm}
\end{figure}

\begin{figure}[t]
\begin{algorithm}[H]
  \caption{TD3 + Self Behavior Cloning (TD3+SelfBC)}
  \label{alg:seperateTD3+SelfBC}
\begin{algorithmic}
  \STATE {\bfseries Hyperparameters:} training steps $N_{\textnormal{SelfBC}}$, and soft update ratio $\tau, \tau_{\textnormal{ref}}$
    
  \STATE {\bfseries Initialize:} Q networks $Q_{\phi_1}, Q_{\phi_2}$ and target Q networks $Q_{\bar\phi_1} Q_{\bar \phi_2}$, policy networks $\pi_\theta$ and target policy networks $\pi_{\bar \theta}$, behavior policy network $\pi_b$, reference policy network $\pi_{\tilde \theta}$

  \STATE {\bfseries Load:} pre-trained Q networks $Q_{\phi_1}, Q_{\phi_2}$ and policy networks $\pi_\theta$

  \STATE // {\bfseries TD3+SelfBC Training Preparation}
  \STATE Update the target network: $\bar{\phi}_{1,2} \leftarrow \phi_{1,2}$, $\bar{\theta} \leftarrow \theta$
  \STATE Update the reference policy network: ${\tilde \theta} \leftarrow \theta$
  
  \STATE // {\bfseries TD3+SelfBC Training}
  \FOR{$t=1$ {\bfseries to} $N_{\textnormal{SelfBC}}$}
    \STATE Sample mini-batch transitions $\left(s, a, r, s^{\prime}\right)$ from $\mathcal{D}$
    \STATE Update $\phi_{1,2}$ by minimizing $J_Q(\phi)$ in Eq.~\eqref{eq:Qloss_TD3+BC}
    \STATE Update $\theta$ by maximizing $J_{\pi}^{\textnormal{TD3+SelfBC}}(\theta)$ in Eq.~\eqref{eq:TD3+SelfBC}
    \STATE Update target networks: \\ \quad
    $\bar{\phi}_{1,2} \leftarrow \tau \phi_{1,2} + (1-\tau) \bar{\phi}_{1,2}$, $\bar{\theta} \leftarrow \tau \theta+(1-\tau) \bar{\theta}$
    \STATE Update reference policy network by Eq.~\eqref{eq:ref_EMA}: \\ \quad
    ${\tilde\theta} \leftarrow \tau_{\textnormal{ref}} \theta + (1 - \tau_{\textnormal{ref}}) \tilde\theta$
  \ENDFOR
\end{algorithmic}
\end{algorithm}
\end{figure}

\begin{figure}[t]
\begin{algorithm}[H]
  \caption{TD3+ESBC}
  \label{alg:TD3+ESBC}
\begin{algorithmic}
  \STATE {\bfseries Hyperparameters:} training steps $N_{\textnormal{SelfBC}}$, soft update ratio $\tau, \tau_{\textnormal{ref}}$, ensemble trainer num $N_{\textnormal{ens}}$
    
  \STATE {\bfseries Initialize:} Q networks $Q_{\phi_1}, Q_{\phi_2}$ and target Q networks $Q_{\bar\phi_1} Q_{\bar \phi_2}$, policy networks $\pi_\theta$ and target policy networks $\pi_{\bar \theta}$, behavior policy network $\pi_b$, reference policy network $\pi_{\tilde \theta}$

  \STATE {\bfseries Load Multiple Trainers:} \\ \quad
  \{pre-trained Q networks $Q_{\phi_1}^i, Q_{\phi_2}^i$, \\ \quad \ \ 
  pre-trained policy networks $\pi_\theta^i | i = 1,2,...,N_{\textnormal{ens}}\}$

  \STATE // {\bfseries Training Preparation}
  \FOR {$i=1$ to $N_{ens}$}
  \STATE Update the target network: $\bar{\phi}_{1,2}^i \leftarrow \phi_{1,2}^i$, $\bar{\theta}^i \leftarrow \theta^i$
  \STATE Update the reference policy network: ${\tilde \theta}^i \leftarrow \theta^i$
  \ENDFOR
  
  \STATE // {\bfseries Training}
  \FOR{$t=1$ {\bfseries to} $N_{\textnormal{SelfBC}}$}
    \STATE Sample mini-batch transitions $\left(s, a, r, s^{\prime}\right)$ from $\mathcal{D}$
    \STATE Compute the average reference action of ensemble policies: \\ \quad 
    $\tilde a = \mathop{Mean}(\{\pi_{\tilde \theta}^i\left(s\right), i=1,2,...,N_{\textnormal{ens}}\})$

    // {\bfseries Train each trainer separately except using the same $\tilde a$}
    \FOR{$i=1$ {\bfseries to} $N_{\textnormal{ens}}$}
    \STATE Update $\phi_{1,2}^i$ by minimizing $J_Q(\phi)$ in Eq.~\eqref{eq:Qloss_TD3+BC}
    \STATE Update $\theta^i$ by maximizing $J_{\pi}^{\textnormal{TD3+ESBC}}(\theta^i)$ in Eq.~\eqref{eq:TD3+ESBC}
    \STATE Update target networks: \\ \quad 
    $\bar{\phi}_{1,2}^i \leftarrow \tau \phi_{1,2}^i + \left(1-\tau\right) \bar{\phi}_{1,2}^i$, $\bar{\theta}^i \leftarrow \tau \theta^i+\left(1-\tau\right) \bar{\theta}^i$
    \STATE Update reference policy network: \\ \quad 
    ${\tilde\theta}^i \leftarrow \tau_{\textnormal{ref}} \theta^i + \left(1 - \tau_{\textnormal{ref}}\right) \tilde\theta^i$
    \ENDFOR
  \ENDFOR
\end{algorithmic}
\end{algorithm}
\end{figure}

\section{Algorithms for TD3+EBC and TD3+SelfBC}\label{appendix:separateAlgorithms}
For completeness, we also  separately present the algorithms for TD3+EBC and TD3+SelfBC as Algorithms~\ref{alg:seperateTD3+EBC} and~\ref{alg:seperateTD3+SelfBC}, respectively. The combined version of the two algorithms was provided in Algorithm~\ref{alg:TD3+SelfBC}. The code will be made available at \url{https://github.com/ShirongLiu/SelfBC}.

\section{Algorithm details of TD3+ESBC}\label{appendix:td3+esbc}
Algorithm~\ref{alg:TD3+ESBC} provides the complete details of TD3+ESBC. In TD3+ESBC, we simultaneously train all the $N_{\textnormal{ens}}$ TD3+SelfBC trainers, where components (such as Q learning, target network updating and reference policy update) of each trainer are independent of others with the exception that the reference action used in the policy objective is the average of the reference policies of all trainers.

\section{Full experimental results for TD3+BC}
\label{Appendix:full_results_alalysis}
In this section, we argue that the degree of policy constraint can effect the performance of TD3+BC, however, relaxing the policy constraint on a static behavior policy does not always improve performance.

\begin{figure*}[thbp!]
    \centering
    \includegraphics[width=1\linewidth]{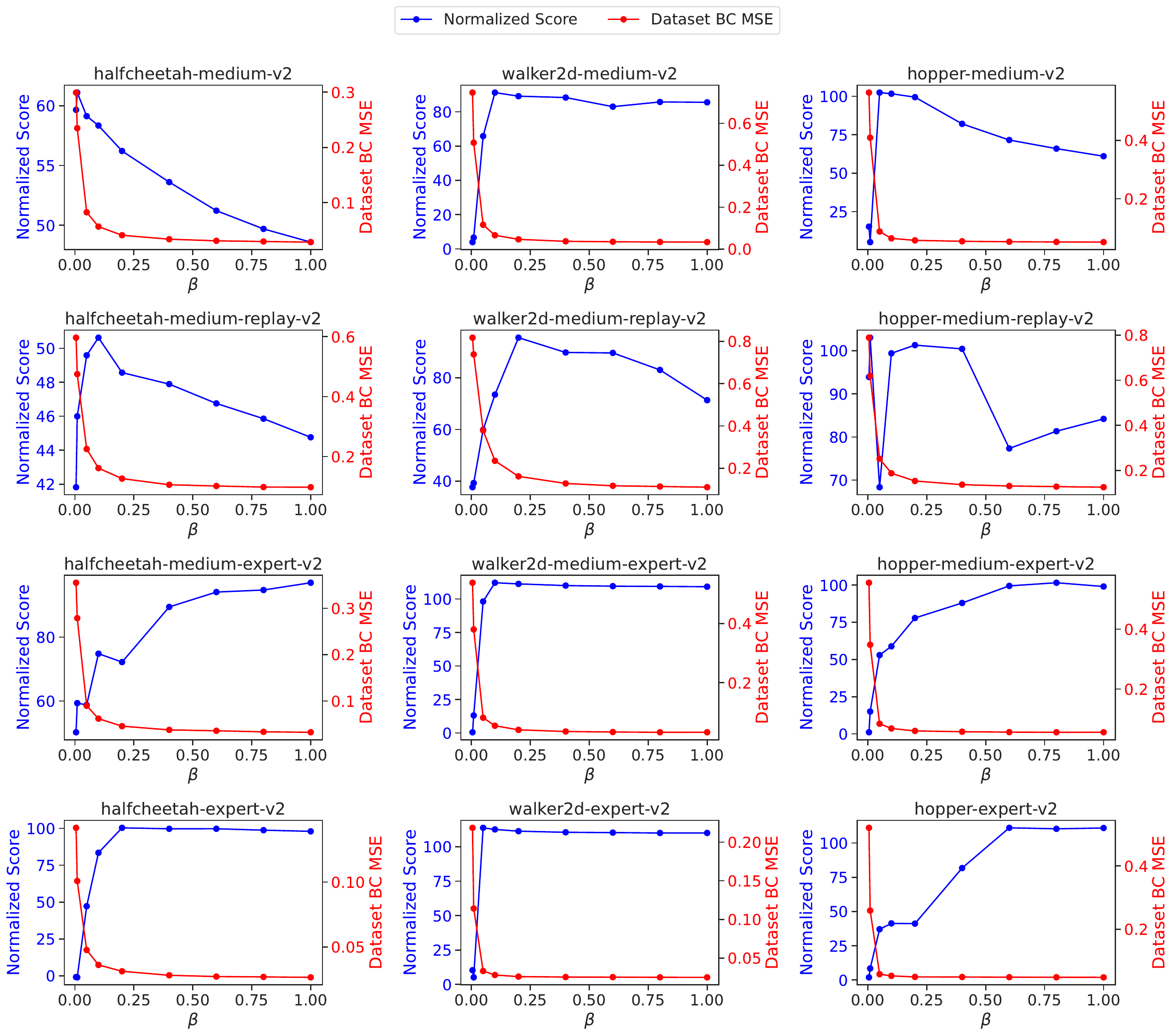}
    \caption{Results, on 12 D4RL\citep{fu2020d4rl} MuJoCo datasets, for TD3+BC trained with different values of the parameter $\beta$ that controls the stringency of the policy constraint. See Fig.~\ref{fig:analys2}(a) caption for additional details.}
    \label{fig:td3bc_beta_score_mse}
\end{figure*}

\begin{figure*}[ht]
    \centering
    \includegraphics[width=1\linewidth]{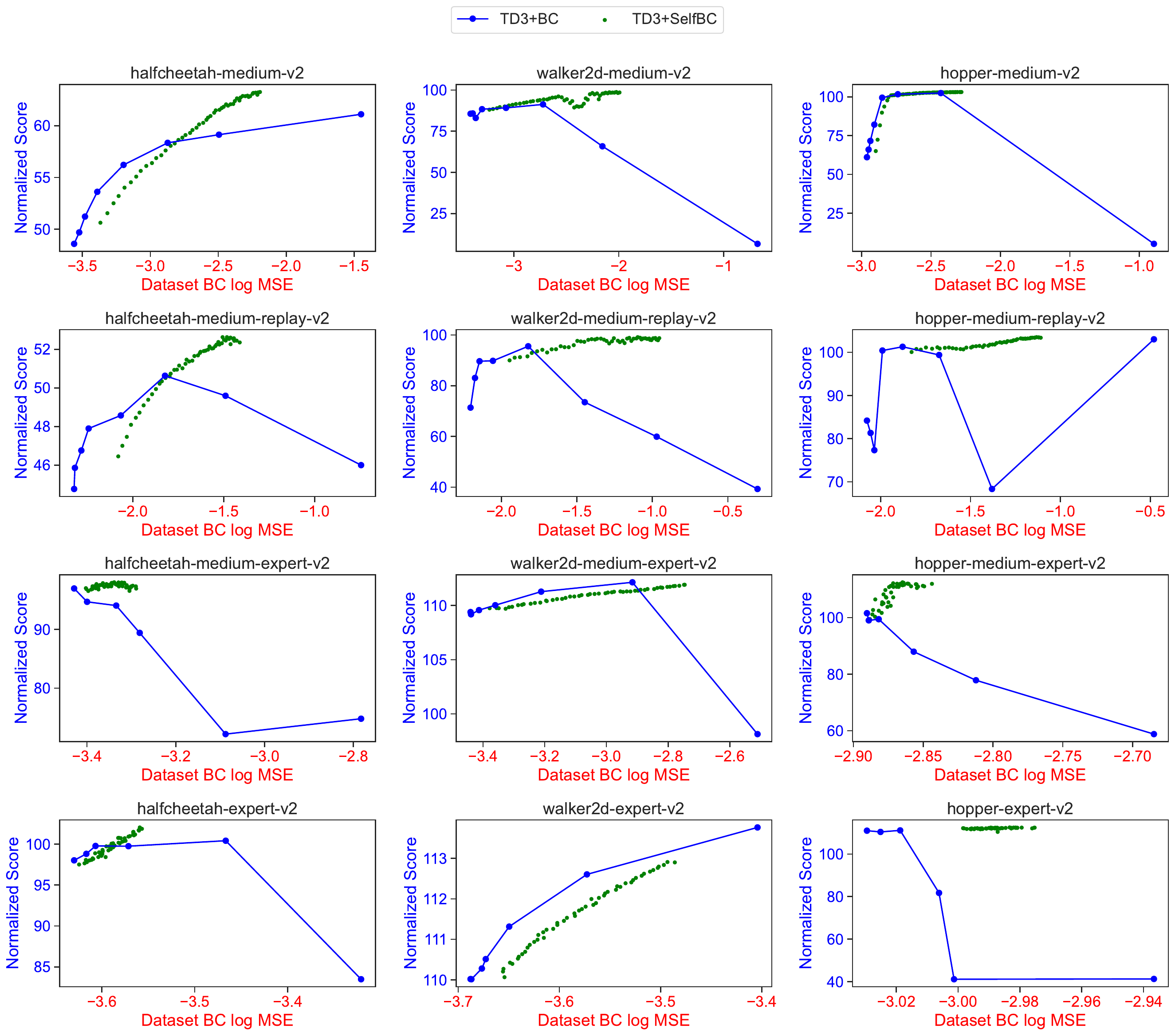}
    \caption{Results, on 12 D4RL\citep{fu2020d4rl} MuJoCo datasets, for TD3+BC (blue) trained with different values of the parameter $\beta$ that controls the stringency of the policy constraint and comparison against the proposed TD3+SelfBC (green). See Fig.~\ref{fig:analys2}(a) caption and the description in the text for additional details.}
    \label{fig:td3bc_mse_score-lineplot}
\end{figure*}

In our objective in Eq.~\eqref{eq:betaaddedTD3nBBObjective} the hyperparameter $\beta$ controls the stringency of the policy constraint. To examine how the stringency of the policy constraint impacts performance, we ran TD3+BC on the 12 offline RL  D4RL \citep{fu2020d4rl} datasets using the alternative $\beta$ values  $1.0, 0.8, 0.6, 0.4, 0.2, 0.1, 0.05, 0.01$, and $0.005$. In Fig. \ref{fig:td3bc_beta_score_mse}, we show the dependence of the normalized score (in blue) and in the dataset BC MSE (in red) on $\beta$. We also provide a fresh perspective for analysis by plotting in Fig. \ref{fig:td3bc_mse_score-lineplot}, the normalized score against the dataset BC log MSE obtained with TD3+BC for different $\beta$ (in blue). Furthermore, we augment Fig. \ref{fig:td3bc_mse_score-lineplot} by plotting, in green, the normalized score and dataset BC log MSE for the proposed TD3+SelfBC algorithm sampled at 50 uniformly spaced epochs during the training process (because we train for 1000 epochs, each green point in Fig. \ref{fig:td3bc_mse_score-lineplot} represents 20 consecutive epochs). To ensure a reasonable scale for visual presentation, select points (those corresponding to $\beta=0.005$ in medium and medium-replay, $\beta=0.005, 0.01$ in walker2d medium-expert and walker2d expert, and $\beta=0.005, 0.01, 0.05$ in other medium-expert and expert) have been omitted from Fig. \ref{fig:td3bc_mse_score-lineplot}.

These results further substantiate the conclusions of Section~\ref{sec:td3bc_analysis} and empirically demonstrate that the proposed TD3+SelfBC approach can learn a better policy after relaxing the constraint, while TD3 + BC may suffer from policy collapse, especially in non-expert datasets such as medium and medium-replay. The experiments demonstrate that TD3+SelfBC achieves significant performance gains, which we can be attributed to the dynamic reference policy in the proposed TD3+SelfBC approach compared with the static reference policy in TD3+BC.

\clearpage

\clearpage

\clearpage

\newpage 
\section{Proof of Theorem \ref{thm:bppo_for_ref}}

We start by introducing some definitions and lemmas. Then we introduce the conservative update process \citep{kakade2002cpi} for our reference policy. Based on these, finally, we provide the proof of Theorem \ref{thm:bppo_for_ref}.
\label{app:proof_of_thm_bppo_for_ref}
\begin{definition}\label{thm:def_bar_a}
Following \citep{zhuang2023behaviorppo}, we define
\begin{eqnarray}
{\bar A}_{\pi,\pi_k}\left(s\right) = \mathbb{E}_{a \sim \pi\left(\cdot|s\right)}\left[ A_{\pi_k}\left(s,a\right) \right]
\end{eqnarray}
by substituting this into Definition \ref{thm:def_J_delta}, we obtain:
\begin{eqnarray}
J_\Delta\left(\pi,\pi_k\right) = \mathbb{E}_{s \sim \rho_{\pi}\left(\cdot\right)}\left[{\bar A}_{\pi, \pi_k}\left(s\right)\right] \\
\widehat{J}_\Delta\left(\pi,\pi_k\right) = \mathbb{E}_{s \sim \rho_D\left(\cdot\right)}\left[{\bar A}_{\pi, \pi_k}\left(s\right)\right]
\end{eqnarray}
\end{definition}

\begin{lemma}\citep{achiam2017constrainedpo}\label{thm:tv_rho_bound_tv_pi}
Given two policies $\pi$ and $\pi_k$, the discrepancy of the state visitation frequencies $\rho_{\pi}, \rho_{\pi_k}$ is bounded by the TV of $\pi, \pi_k$.
\begin{eqnarray}
\left\| \rho_{\pi} - \rho_{\pi_k} \right\|_1
\leq 
\frac{2\gamma}{1-\gamma} \mathbb{E}_{s \sim \rho_{\pi_k}}\left[{D_{TV}}\left(\pi \| \pi_k\right)\left[s\right]\right]
\end{eqnarray}
where $D_{TV}\left(p \|  q\right)\left[s\right]$ is the TV of two action distributions $p, q$ at $s$, defined as 
\begin{eqnarray}
D_{TV}\left(p \|  q\right)\left[s\right] = \frac{1}{2} \sum_{a} | p(a|s) - q(a|s) |
\end{eqnarray}
\end{lemma}

\begin{lemma}\citep{zhuang2023behaviorppo}\label{thm:bppo_origin}
The variation in the performance difference between two policies $\pi$ and $\pi_k$ influenced by the offline approximation can be bounded as
\begin{eqnarray}
&&\left| J_\Delta\left(\pi,\pi_k\right) - \widehat{J}_\Delta\left(\pi,\pi_k\right) \right| \nonumber \\
&=&
\left| 
    \mathbb{E}_{s \sim \rho_{\pi}\left(\cdot\right)}\left[{\bar A}_{\pi, \pi_k}\left(s\right)\right] - \mathbb{E}_{s \sim \rho_D\left(\cdot\right)}\left[{\bar A}_{\pi, \pi_k}\left(s\right)\right] 
\right| \nonumber \\
&\leq&
\left|
    \mathbb{E}_{s \sim \rho_{\pi}\left(\cdot\right)}\left[
    {\bar A}_{\pi, \pi_k}\left(s\right)\right] 
    - \mathbb{E}_{s \sim \rho_{\pi_k}\left(\cdot\right)}\left[{\bar A}_{\pi, \pi_k}\left(s\right)\right] 
\right| \nonumber \\
&+& 
\left|
    \mathbb{E}_{s \sim \rho_{\pi_k}\left(\cdot\right)}\left[{\bar A}_{\pi, \pi_k}\left(s\right)\right]
    - \mathbb{E}_{s \sim \rho_{\pi_b}\left(\cdot\right)}\left[{\bar A}_{\pi, \pi_k}\left(s\right)\right] 
\right| \nonumber \\
&+& 
\left|
    \mathbb{E}_{s \sim \rho_{\pi_b}\left(\cdot\right)}\left[{\bar A}_{\pi, \pi_k}\left(s\right)\right]
    - \mathbb{E}_{s \sim \rho_D\left(\cdot\right)}\left[{\bar A}_{\pi, \pi_k}\left(s\right)\right] 
\right| \nonumber
\end{eqnarray}
where $\pi_b$ is the behavior policy learned from the offline dataset $\mathcal{D}$.
\end{lemma}

\textbf{Conservative update of reference policy} We then focus on the specific reference policy update process mentioned in Definition \ref{thm:def_pi}, which denotes $\pi$ as the new reference policy, $\pi_k$ as the reference policy of the k-th iteration, and $\pi'$ as the new learned policy.
\begin{eqnarray}\label{eq:cpi_update}
    \pi\left(\cdot|s\right) = \left(1-\kappa\right) \pi_k\left(\cdot|s\right) + \kappa \pi'\left(\cdot|s\right)
\end{eqnarray}

We substitute Eq.~\eqref{eq:cpi_update} to Definition \ref{thm:def_bar_a} and obtain:
\begin{eqnarray}
&&{\bar A}_{\pi, \pi_k}\left(s\right) \nonumber \\
&=&\mathbb{E}_{a \sim \pi\left(\cdot|s\right)}\left[ A_{\pi_k}\left(s,a\right) \right] \nonumber \\
&=&\mathbb{E}_{a \sim \left(1-\kappa\right) \pi_k\left(\cdot|s\right) + \kappa \pi'\left(\cdot|s\right)}\left[ A_{\pi_k}\left(s,a\right) \right] \nonumber \\
&=&(1-\kappa)\mathbb{E}_{a \sim \pi_k\left(\cdot|s\right)}\left[A_{\pi_k}\left(s,a\right) \right]
+\kappa \mathbb{E}_{a \sim \pi'\left(\cdot|s\right)}\left[A_{\pi_k}\left(s,a\right)\right]
\nonumber \\
&=&\kappa \mathbb{E}_{a \sim \pi'\left(\cdot|s\right)}\left[A_{\pi_k}\left(s,a\right)\right] \nonumber \\
&=&\kappa {\bar A}_{\pi', \pi_k}\left(s\right)
\end{eqnarray}
where the final step uses Definition \ref{thm:def_bar_a}, and the penultimate step uses:
\begin{eqnarray}\label{eq:def_bar_a'}
&&\mathbb{E}_{a \sim \pi_k\left(\cdot|s\right)}\left[A_{\pi_k}\left(s,a\right) \right] \nonumber \\
&=&\mathbb{E}_{a \sim \pi_k\left(\cdot|s\right)}\left[Q_{\pi_k}\left(s,a\right) - V_{\pi_k}\left(s\right) \right] = 0
\end{eqnarray}

We substitute Eq.~\eqref{eq:cpi_update} to $D_{TV}\left(\pi, \pi_k\right)$ and obtain:
\begin{align}\label{eq:def_TV_pi'}
D_{TV}\left(\pi, \pi_k\right) &= \frac{1}{2} \left\|\pi - \pi_k \right\|_1 = \frac{1}{2} \left\|\left(1-\kappa\right)\pi_k + \kappa \pi' - \pi_k \right\|_1 \nonumber\\
&= \frac{1}{2} \kappa \left\| \pi' - \pi_k \right\|_1 = \kappa D_{TV}\left(\pi' \| \pi_k\right)
\end{align}

We summarize Eq.~\eqref{eq:def_bar_a'} and Eq.~\eqref{eq:def_TV_pi'} as the following conclusion:
\begin{corollary}\label{thm:bar_a_ref}
When applying conservative update for the reference policy in Eq.~\eqref{eq:cpi_update}, we have:
\begin{eqnarray}
{\bar A}_{\pi, \pi_k}\left(s\right)&=&\kappa {\bar A}_{\pi', \pi_k}\left(s\right) \\
D_{TV}\left(\pi, \pi_k\right) &=& \kappa D_{TV}\left(\pi' \| \pi_k\right)
\end{eqnarray}
\end{corollary}

We then rewrite Lemma \ref{thm:bppo_origin} by applying the conservative policy update in Eq.~\eqref{eq:cpi_update}. In the following development, we separately consider the three terms in the right hand side of Lemma \ref{thm:bppo_origin}:

The first term in the right hand side of Lemma \ref{thm:bppo_origin} is:
\begin{eqnarray}
\left|
    \mathbb{E}_{s \sim \rho_{\pi}\left(\cdot\right)}\left[
    {\bar A}_{\pi, \pi_k}\left(s\right)\right] 
    - \mathbb{E}_{s \sim \rho_{\pi_k}\left(\cdot\right)}\left[{\bar A}_{\pi, \pi_k}\left(s\right)\right] 
\right|
\end{eqnarray}

The second term in the right hand side of Lemma \ref{thm:bppo_origin} is: 
\begin{eqnarray}
\left|
    \mathbb{E}_{s \sim \rho_{\pi_k}\left(\cdot\right)}\left[{\bar A}_{\pi, \pi_k}\left(s\right)\right]
    - \mathbb{E}_{s \sim \rho_{\pi_b}\left(\cdot\right)}\left[{\bar A}_{\pi, \pi_k}\left(s\right)\right] 
\right|
\end{eqnarray}

The third term in the right hand side of Lemma \ref{thm:bppo_origin} is: 
\begin{eqnarray}
\left|
    \mathbb{E}_{s \sim \rho_{\pi_b}\left(\cdot\right)}\left[{\bar A}_{\pi, \pi_k}\left(s\right)\right]
    - \mathbb{E}_{s \sim \rho_D\left(\cdot\right)}\left[{\bar A}_{\pi, \pi_k}\left(s\right)\right] 
\right|
\end{eqnarray}

\textbf{The first term can be derived as follows:}
\begin{eqnarray}
&&
\left|
    \mathbb{E}_{s \sim \rho_{\pi}\left(\cdot\right)}\left[
    {\bar A}_{\pi, \pi_k}\left(s, a\right)\right] 
    - \mathbb{E}_{s \sim \rho_{\pi_k}\left(\cdot\right)}\left[{\bar A}_{\pi, \pi_k}\left(s, a\right)\right] 
\right| \nonumber \\
&=&\kappa  \left|
    \mathbb{E}_{s \sim \rho_{\pi}\left(\cdot\right)}\left[
    {\bar A}_{\pi', \pi_k}\left(s\right)\right] 
    - \mathbb{E}_{s \sim \rho_{\pi_k}\left(\cdot\right)}\left[{\bar A}_{\pi', \pi_k}\left(s\right)\right] 
\right| \nonumber \\
&\leq& \kappa \left\|\rho_\pi - \rho_{\pi_k}\right\|_{1}  \left\| {\bar A}_{\pi', \pi_k}(s) \right\|_{\infty} \nonumber \\
&\leq& \frac{2\kappa\gamma}{1-\gamma} \mathbb{E}_{s \sim \rho_{\pi_k}}\left[{D_{TV}}\left(\pi \| \pi_k\right)\left[s\right]\right] 
\max_s \left| {\bar A}_{\pi', \pi_k}\left(s\right) \right| \nonumber \\
&\leq& \frac{2 \kappa^2 \gamma}{1-\gamma} \mathbb{E}_{s \sim \rho_{\pi_k}}\left[{D_{TV}}\left(\pi' \| \pi_k\right)\left[s\right]\right] 
\max_s \left| {\bar A}_{\pi', \pi_k}\left(s\right) \right| \nonumber \\
\end{eqnarray}
where the second line uses Conclusion \ref{thm:bar_a_ref}, the third line uses Hölder’s inequality, the second line from the bottom uses Lemma \ref{thm:tv_rho_bound_tv_pi}, and the bottom line uses Conclusion \ref{thm:bar_a_ref}.

\textbf{The second term can be derived as follows:}
\begin{eqnarray}\label{eq:second_term_result}
&&
\left|
\mathbb{E}_{s \sim \rho_{\pi_k}\left(\cdot\right)}\left[{\bar A}_{\pi, \pi_k}\left(s\right)\right]
- \mathbb{E}_{s \sim \rho_{\pi_b}\left(\cdot\right)}\left[{\bar A}_{\pi, \pi_k}\left(s\right)\right] 
\right| \nonumber \\
&=& \kappa \left|
\mathbb{E}_{s \sim \rho_{\pi_k}\left(\cdot\right)}\left[{\bar A}_{\pi', \pi_k}\left(s\right)\right]
- \mathbb{E}_{s \sim \rho_{\pi_b}\left(\cdot\right)}\left[{\bar A}_{\pi', \pi_k}\left(s\right)\right] 
\right| \nonumber \\
&\leq& \kappa \left\|\rho_{\pi_k} - \rho_{\pi_b}\right\|_{1}  \left\| {\bar A}_{\pi', \pi_k}\left(s\right) \right\|_{\infty} \nonumber \\
&\leq& \frac{2\kappa\gamma}{1-\gamma} \mathbb{E}_{s \sim \rho_{\pi_b}}\left[{D_{TV}}\left(\pi_k \| \pi_b\right)\left[s\right]\right] 
\max_s \left| {\bar A}_{\pi', \pi_k}\left(s\right) \right| \nonumber \\
\end{eqnarray}
where the second line uses Conclusion \ref{thm:bar_a_ref}, the third line uses Hölder’s inequality, and the bottom line uses Lemma 4.

\textbf{The third term can be derived as follows:}
\begin{eqnarray}
&&
\left|
    \mathbb{E}_{s \sim \rho_{\pi_b}(\cdot)}\left[{\bar A}_{\pi, \pi_k}\left(s\right)\right]
    - \mathbb{E}_{s \sim \rho_{\mathcal{D}}\left(\cdot\right)}\left[{\bar A}_{\pi, \pi_k}\left(s\right)\right] 
\right| \nonumber \\
&=& \kappa \left|
    \mathbb{E}_{s \sim \rho_{\pi_b}\left(\cdot\right)}\left[{\bar A}_{\pi', \pi_k}\left(s\right)\right]
    - \mathbb{E}_{s \sim \rho_{\mathcal{D}}\left(\cdot\right)}\left[{\bar A}_{\pi', \pi_k}\left(s\right)\right] 
\right| \nonumber \\
&\leq& \frac{2\kappa\gamma}{1-\gamma} \mathbb{E}_{s \sim \rho_{\mathcal{D}}}\left[{D_{TV}}\left(\pi_b \| \pi_{\mathcal{D}}\right)\left[s\right]\right] 
\max_s \left| {\bar A}_{\pi', \pi_k}\left(s\right) \right| \nonumber \\
&=& \frac{\kappa\gamma}{1-\gamma} \mathbb{E}_{s \sim \rho_{\mathcal{D}}} \left(1-\pi_b\left(a|s\right)\right)
\max_s \left| {\bar A}_{\pi', \pi_k}\left(s\right) \right|
\end{eqnarray}
where the second and the third line are derived similarly to the above two terms, and the last line uses the following equation from \citep{zhuang2023behaviorppo}:
\begin{eqnarray}
{D_{TV}}\left(\pi_b \| \pi_{\mathcal{D}}\right)\left[s\right] = \frac{1}{2}\left(1-\pi_b\left(a|s\right)\right)
\end{eqnarray}

Using the result from \citep{zhuang2023behaviorppo}, we bound the common term $\max_s \left| {\bar A}_{\pi', \pi_k}\left(s\right) \right|$ in the preceding three derivations as
\begin{eqnarray}
&&\max_s \left| 
    {\bar A}_{\pi', \pi_k}\left(s\right) 
\right| \nonumber \\
&=&
\max_s \left|
    \mathbb{E}_{a \sim \pi'\left(\cdot|s\right)}\left[ A_{\pi_k}\left(s,a\right) \right]
    - \mathbb{E}_{a \sim \pi_k\left(\cdot|s\right)}\left[ A_{\pi_k}\left(s,a\right) \right]
\right| \nonumber \\
&\leq&
\max_s \left|
    \left\| \pi'\left(\cdot|s\right) - \pi_k\left(\cdot|s\right) \right\|_1
    \left\| A_{\pi_k}\left(s,a\right) \right\|_{\infty}
\right| \nonumber \\
&=&
\max_s \left|
    2D_{TV}\left(\pi' \| \pi_k\right)\left[s\right] \cdot 
    \max_a A_{\pi_k}\left(s,a\right)
\right| \nonumber \\
&\leq&
2 \max_s 
D_{TV}\left(\pi' \| \pi_k\right)\left[s\right] \cdot 
\max_{s,a} \left|A_{\pi_k}\left(s,a\right)\right| \nonumber \\
&\triangleq&\mathbb{A}_{\pi', \pi_k} 
\end{eqnarray}

\textbf{Distribution shift:}
For the derivation result of the second term in Eq.~\eqref{eq:second_term_result}, we obtain a further bound based on Hölder’s inequality:
\begin{eqnarray}
\left\| \rho_\pi - \rho_{\pi_b} \right\|_1
\leq 
\left\| \rho_\pi - \rho_{\pi_k} \right\|_1 + \left\| \rho_{\pi_k} - \rho_{\pi_b} \right\|_1
\end{eqnarray}
where the second term $\left\| \rho_{\pi_k} - \rho_{\pi_b} \right\|_1$ is independent of the new learned policy $\pi'$, and the first term can be further bounded as:
\begin{eqnarray}
&&\left\| \rho_\pi - \rho_{\pi_k} \right\|_1 \nonumber \\
&\leq& \frac{2\gamma}{1-\gamma} \mathbb{E}_{s \sim \rho_{\pi_k}}\left[{D_{TV}}(\pi \| \pi_k)\left[s\right]\right] \nonumber \\
&\leq& \frac{2\kappa\gamma}{1-\gamma} \mathbb{E}_{s \sim \rho_{\pi_k}}\left[{D_{TV}}(\pi' \| \pi_k)\left[s\right]\right]
\end{eqnarray}
Thus, to bound $\left\| \rho_\pi - \rho_{\pi_b} \right\|_1$, the unnormalized state visitation frequencies shift between the learned behavior policy $\pi_b$ and the new reference policy $\pi$, we only need to bound ${D_{TV}}\left(\pi' \| \pi_k\right)\left[s\right]$, the distribution shift between the new learned policy $\pi'$ and the old reference policy $\pi_k$.

We finish our proof by combining the above three terms and the definition of $\mathbb{A}_{\pi',\pi_k}$. The variation in performance difference of two consecutive reference policies $\pi, \pi_k$ influenced by offline approximation is bounded as:
\begin{eqnarray}\label{eq:bound_combine}
&& \left| J_{\Delta}\left(\pi, \pi_k\right) - \widehat{J}_{\Delta}\left(\pi, \pi_k\right) \right| 
\nonumber \\
&\leq&
\frac{2\gamma\kappa^2}{1-\gamma}
\mathbb{A}_{\pi',\pi_k}
\mathbb{E}_{s\sim\rho_{\pi_k}} \left[ D_{TV}\left(\pi' \| \pi_k\right)\left[s\right] \right]
\nonumber \\
&+& 
\frac{2\gamma\kappa}{1-\gamma}
\mathbb{A}_{\pi',\pi_k}
\mathbb{E}_{s\sim\rho_{\pi_b}} \left[ D_{TV}\left(\pi_k \| \pi_b\right)\left[s\right] \right] 
\nonumber \\
&+& 
\frac{\gamma\kappa}{1-\gamma}
\mathbb{A}_{\pi',\pi_k}
\mathbb{E}_{s\sim\rho_{\mathcal{D}}} \left[ 1-\pi_b\left(a|s\right) \right]
\end{eqnarray}
where $\mathbb{A}_{\pi',\pi_k}$ is defined as 
\begin{eqnarray}
\mathbb{A}_{\pi',\pi_k} = 2 \max_{s,a} \left| A_{\pi_k\left(s,a\right)} \right| \max_s D_{TV}\left(\pi' \| \pi_k\right)\left[s\right]
\end{eqnarray}

The bound in Eq.~\eqref{eq:bound_combine} directly implies the following bound:
\begin{eqnarray}\label{eq:thm_appendix_rewritten}
J_{\Delta}\left(\pi, \pi_k\right)
&\geq&
\widehat{J}_{\Delta}\left(\pi, \pi_k\right) 
\nonumber \\
&-& 
\frac{2\gamma\kappa^2}{1-\gamma}
\mathbb{A}_{\pi',\pi_k}
\mathbb{E}_{s\sim\rho_{\pi_k}} \left[ D_{TV}\left(\pi' \| \pi_k\right)\left[s\right] \right]
\nonumber \\
&-& 
\frac{2\gamma\kappa}{1-\gamma}
\mathbb{A}_{\pi',\pi_k}
\mathbb{E}_{s\sim\rho_{\pi_b}} \left[ D_{TV}\left(\pi_k \| \pi_b\right)\left[s\right] \right] 
\nonumber \\
&-& 
\frac{\gamma\kappa}{1-\gamma}
\mathbb{A}_{\pi',\pi_k}
\mathbb{E}_{s\sim\rho_{\mathcal{D}}} \left[ 1-\pi_b\left(a|s\right) \right]
\end{eqnarray}
where $\mathbb{A}_{\pi',\pi_k}$ is as defined above.
\paragraph{Monotonic improvement guarantee}
In Eq.~\eqref{eq:thm_appendix_rewritten}, the L.H.S is the improvement in expected discounted reward for policy $\pi$ over policy $\pi_k$. For guaranteeing monotonic improvement, it suffices if this term is positive ($J_{\Delta}\left(\pi, \pi_k\right) > 0$), we do not necessarily need to maximize $J_{\Delta}\left(\pi, \pi_k\right)$. Thus, if the lower bound in the R.H.S of Eq.~\eqref{eq:thm_appendix_rewritten} is maximized while maintaining a positive value, a corresponding monotonic improvement is guaranteed.

\paragraph{Adaptive rate $\kappa$} In the original CPI \cite{kakade2002cpi} paper, the mixture rate $\kappa$ is adaptively chosen to guarantee the monotonic improvement. Deep CPI \cite{vieillard2020deepcpi} approximately estimates the rate but only considers the discrete action space, where the greedy policy $\pi'$ can be obtained directly, so the performance difference of the greedy policy and the current policy can be calculated directly.

In the offline-learning continuous action space setting that we consider in this paper, the estimation of $\kappa$ is complicated by the following differences:
\begin{itemize}
    \item the greedy policy $\pi'$ can not be directly obtained and can only be approximated by policy optimization using the TD3+SelfBC objective \eqref{eq:TD3+SelfBC}.
    \item we consider the performance difference in the offline setting, where the state distribution follows the dataset behavior policy.
\end{itemize}

For our experiments, we empirically choose a really small value (5e-5 or 5e-6) for the mixing rate $\kappa$ during training, which achieves nearly monotonic improvement in practice but does not offer absolute guarantees. We leave a principled determination of the mixture rate $\kappa$ to future work.


\section{Baselines}\label{app:baselines}
We compare our proposed methods with several competitive policy constraint methods: including TD3+BC \citep{fujimoto2021td3bc}, ReBRAC \citep{tarasov2024rebrac}, wPC \citep{peng2023wPC}, CPI-RE \citep{hu2023cpi_re}, and we also compare our methods with two value conservative methods SAC-RND \citep{nikulin2023sac-rnd} and EDAC \citep{an2021edac}. We simply introduce these algorithms:
\begin{itemize}
    \item TD3+BC: an approach for offline RL that incorporates behavior cloning (BC) as a policy regularization within TD3.
    \item ReBRAC: a method incorporates BC within TD3, and proposes several design elements such as larger batches, layer norm for the critic networks, critic penalty, deep networks, and hyperparameter tuning.
    \item wPC: a method that improves TD3+BC by modifying the original BC term to the weighted BC term. By assigning identifier weights, it have the capacity of selecting high-quality dataset samples, which improves the quality of the policy constraint.
    \item CPI-RE: a method that improves TD3+BC by progressively updating the reference as a copy of the current learned policy. It differs from the proposed TD3+SelfBC in three crucial aspects: (1) The TD3+SelfBC reference policy has a more conservative soft-update process via EMA, instead of the hard-update process in CPI-RE. (2) the policy constraint in CPI-RE contains both the original BC term and their proposed constraint term; in contrast, the policy constraint in TD3+SelfBC only contains the SelfBC term. (3) For the hyperparameters, CPI-RE involves the coefficient of policy constraint, in contrast, we do not involve any additional coefficients, but only utilize $\tau_{\textnormal{ref}}$ to control the update speed of our reference policy.      
    \item SAC-RND: an anti-exploration method for offline RL, which employs random network distillation (RND) as regularization for both policy learning and value learning.
    \item EDAC: an uncertainty-based method for offline RL, which implicitly penalizes the value learning by the uncertainty via ensemble minimization.
\end{itemize}

\section{Hyperparameters}\label{app:hyperparameters}
The hyperparameter values used for TD3+EBC and TD3+SelfBC are listed in Table \ref{tab:hyperparameters-td3bc} and Table \ref{tab:hyperparameters-special}. Table \ref{tab:hyperparameters-td3bc} enumerates the TD3+BC common hyperparameters, which are also used by TD3+SelfBC and in the pretraining of BC, TD3+BC, TD3+EBC. Table \ref{tab:hyperparameters-special} lists the values of other hyperparameters used by pretraining algorithms and TD3+SelfBC.

These hyperparameters are empirically chosen without intricate tuning for each dataset, largely following TD3+BC (e.g. for the coefficient $\alpha$). We expect $\tau_{\textnormal{ref}}$ to be smaller than $\tau$, the update ratio of the target Q network, so we set $\tau_{\textnormal{ref}} = (0.01 \ \textnormal{or}\ 0.001) * \tau$. For expert datasets, we further reduce $\alpha$ and $\tau_{\textnormal{ref}}$ to achieve a more stable but slower policy learning.
For comparison methods, wPC simply reduces $\alpha$ for ``medium-expert'' datasets; ReBRAC, CPI-RE, SAC-RND tune hyperparameters for each dataset; EDAC maintains hyperparameter consistency within each environment. In general, to be fair to other methods in comparisons, we maintain consistency at the ``dataset level''.

\begin{table}[H]
\centering
\caption{TD3+BC common hyperparameters. 
}
\label{tab:hyperparameters-td3bc}
\begin{tabular}{c|c}
\toprule
Hyperparameters & Value \\
\midrule
batch size & 256 \\
discount $\gamma$ & 0.99 \\
policy noise & 0.2 \\
noise clip & 0.5 \\
policy update frequency & 2 \\
state normalization & True \\
Q normalization & True \\
critic Layer Normalization & True \\
\bottomrule
\end{tabular}
\end{table}

\begin{table}[H]
\centering
\caption{Special Hyperparameters for Pretraining Algorithms and TD3+SelfBC}
\label{tab:hyperparameters-special}
\begin{tabular}{c|c|c|c|c}
\toprule
\multirow{2}{*}{Dataset} & \multicolumn{2}{c|}{Pretrain Algorithms} & \multicolumn{2}{c}{TD3+SelfBC} \\
\cmidrule{2-3} \cmidrule{4-5}
& $\alpha_{\text{pretrain}}$ & $N_{\text{pretrain}}$ & $\alpha_{\text{self}}$ & $\tau_{\text{ref}}$ \\
\midrule
HC-m & \multirow{6}{*}{2.5} & \multirow{6}{*}{2e5} & \multirow{6}{*}{2.5} & \multirow{6}{*}{5e-5} \\
HC-mr &  &  &  &  \\
W-m &  &  &  &  \\
W-mr &  &  &  &  \\
Hop-m &  &  &  &  \\
Hop-mr &  &  &  &  \\
\midrule
HC-me & \multirow{6}{*}{1.0} & 1e6 & \multirow{6}{*}{1.0} & 5e-6 \\
HC-e &  & 1e6 &  & 5e-6 \\
W-me &  & 2e5 &  & 5e-5 \\
W-e &  & 2e5 &  & 5e-5 \\
Hop-me &  & 1e6 &  & 5e-6 \\
Hop-e &  & 1e6 &  & 5e-6 \\
\bottomrule
\end{tabular}
\end{table}

\section{Computation cost}
We run all our experiments on a single 1080ti GPU. We compare the training time and GPU memory usage of the pretraining algorithms (BC, TD3+BC, TD3+EBC) and our proposed algorithms (TD3+SelfBC, TD3+ESBC). 

\paragraph{Single learning loop}
As completely outlined in Algorithm~\ref{alg:TD3+SelfBC}, TD3+SelfBC contains three phases: (1) load pre-trained networks (by BC, TD3+BC or TD3+EBC). (2) training preparation. (3) TD3+SelfBC training. So TD3+SelfBC involves a double learning loop process (both the first and the third phase need training). To simplifying the presentation/discussion, we only consider the training time of the third part ($N_{\textnormal{SelfBC}}$ training steps), which we call  ``single learning loop'' and report this in Table~\ref{tab:computation_cost_single_loop}.
Note that for TD3+EBC, we also only consider the $N_{\textnormal{EBC}}$ training steps and ignore the BC training part, and we consider TD3+ESBC in the same way as TD3+SelfBC.

\begin{table}[h]
    \centering
    \caption{Time and GPU memory usage (single learning loop)}
    \label{tab:computation_cost_single_loop}
    \begin{tabular}{c|c|c}
    \toprule
    Algorithms & 1000 epochs time & GPU Memory \\
    \midrule
    BC & 40m & \multirow{5}{*}{1.2GB} \\
    TD3+BC & 1h40m & \\
    TD3+EBC & 2h & \\
    TD3+SelfBC & 2h & \\
    TD3+ESBC & 10h & \\
    \bottomrule
    \end{tabular}
\end{table}

In Table~\ref{tab:computation_cost_single_loop}, the prolonged time for TD3+ESBC is because our implementation trains the $N_{\textnormal{ens}}=5$ trainers sequentially in the ``for loop'' in Algorithm~\ref{alg:TD3+ESBC}. Parallel training, realized through vectorization techniques used in EDAC \cite{an2021edac}, can significantly reduce the time for TD3+ESBC.

\paragraph{Double learning loop}
When considering the whole training process for TD3+SelfBC, the TD3+EBC is trained first, and the TD3+SelfBC is subsequently trained. Compared with the single learning loop, the GPU memory usage is not increased because there is no additional network. Although the pretraining process consumes additional time, the total time is still acceptable. For example, for ``medium'' and ``medium-replay'' datasets, TD3+EBC is trained for 200 epochs (0.4h), after which TD3+SelfBC is trained for 1000 epochs (2h). The total time is 2.4h. For ``medium-expert'' and ``expert'' datasets, both are trained for 1000 epochs, and total time is 4h.

\paragraph{Comparison with other algorithms}
SAC-RND and EDAC run 3000 epochs for better convergence, our total epochs are lower.
Policy constraint methods usually run 1000 epochs because of fast convergence. For example, we rerun TD3+BC for 2000 epochs on 4 seeds, the performance (HC-m 48.9, HC-mr 45.6, W-m 85.6, W-mr 81.6, Hop-m 61.5, Hop-mr 64.5) does not show significant difference from 1000 epochs (even 500 epochs), probably because of the use of a static constraint.

\section{Additional results for TD3+SelfBC (select)}
Recall that in Table~\ref{tab:d4rl-mujoco}, we evaluate the TD3+SelfBC (select) version, which loads the manually selected pre-trained results for best performance of TD3+SelfBC. In this section, we additionally evaluate the ``select worst'' version, which loads another manually selected pre-trained results, but for worst performance of TD3+SelfBC. We show the results in Table~\ref{tab:select_worst}, and compare with the two variants in Table~\ref{tab:d4rl-mujoco}, note that we rename the ``TD3+SelfBC (select)'' in Table~\ref{tab:d4rl-mujoco} to ``select (best)'' in Table~\ref{tab:select_worst}.

\begin{table}[ht]
\centering
\setlength{\tabcolsep}{2.2pt}
\caption{Results of the TD3+SelfBC (select worst) version.
}
\label{tab:select_worst}
\begin{tabular}{c|cc|c}
\toprule
Dataset & TD3+SelfBC & select (best) & select (worst) \\
\midrule
HC-m & 61.0 $\pm$ 1.1 & \textbf{63.3} $\pm$ 0.2 & 61.1 $\pm$ 0.9 \\
HC-mr & 52.3 $\pm$ 0.5 & \textbf{52.3} $\pm$ 0.6 & 51.3 $\pm$ 0.7 \\
\midrule
W-m & 85.7 $\pm$ 17.9 & \textbf{99.1} $\pm$ 0.6 & 60.5 $\pm$ 37.2 \\
W-mr & 95.4 $\pm$ 4.8 & \textbf{99.4} $\pm$ 1.4 & 92.2 $\pm$ 6.4 \\
\midrule
Hop-m & 102.9 $\pm$ 0.1 & \textbf{103.2} $\pm$ 0.2 & 103.1 $\pm$ 0.2 \\
Hop-mr & 101.7 $\pm$ 1.6 & \textbf{103.6} $\pm$ 0.4 & 103.5 $\pm$ 0.4 \\
\midrule
Avg on m and m-r & 83.2 & \textbf{86.8} & 78.6 \\
\bottomrule
\end{tabular}
\end{table}

\end{document}